\pdfoutput=1

\documentclass[11pt]{article}

\usepackage{EMNLP2023}

\usepackage{times}
\usepackage{latexsym}

\usepackage[T1]{fontenc}

\usepackage[utf8]{inputenc}

\usepackage{microtype}

\usepackage{inconsolata}
\usepackage{amsmath}
\usepackage{graphicx}
\usepackage{booktabs}
\usepackage{bbm}
\usepackage{subcaption}
\usepackage{tabularx,ragged2e}
\usepackage{multicol,multirow}
\usepackage{enumitem}
\usepackage{soul}
\usepackage{arydshln}
\usepackage{bm}
\usepackage{pifont}
\newcommand{\xmark}{\ding{55}}%
\newcommand{\cmark}{\ding{51}}%

\definecolor{carminered}{rgb}{1.0, 0.0, 0.22}
\definecolor{coralred}{rgb}{0.93, 0, 0}

\title{Detecting and Mitigating Hallucinations via Self-Critiquing}
\title{Detecting and Mitigating Hallucinations of Text Generation Models: A Systematic Approach}
\title{\textit{A Stitch in Time Saves Nine}: Actively Detecting and Mitigating Hallucinations of Large Language Models}
\title{\textit{A Stitch in Time Saves Nine}: Detecting and Mitigating LLMs' Hallucinations Leveraging Logit Outputs and Web Search}
\title{\textit{A Stitch in Time Saves Nine}: Detecting and Mitigating Large Language Models' Hallucinations Leveraging Logit Outputs and Web Search}
\title{\textit{A Stitch in Time Saves Nine}: Detecting and Mitigating Hallucinations of LLMs by Validating Low-Confidence Generation}

\author{Neeraj Varshney$^{\diamondsuit}$\thanks{~~Work done during  internship at Tencent AI Lab} 
\hspace{9pt}
Wenlin Yao$^{\clubsuit}$ \hspace{9pt} 
Hongming Zhang$^{\clubsuit}$ \\ 
\textbf{Jianshu Chen}$^{\clubsuit}$ \hspace{9pt} 
\textbf{Dong Yu}$^{\clubsuit}$  \\
$^\diamondsuit$Arizona State University \; $^\clubsuit$ Tencent AI Lab, Bellevue, WA \\
$^{\diamondsuit}$\texttt{nvarshn2@asu.edu} \\
$^{\clubsuit}$\texttt{\{wenlinyao, hongmzhang, jianshuchen, dyu\}@global.tencent.com}
}

\begin{document}

\maketitle
\begin{abstract}
Recently developed large language models have achieved remarkable success in generating fluent and coherent text. However, these models often tend to `hallucinate' which critically hampers their reliability.
In this work, we address this crucial problem and propose an approach that actively detects and mitigates hallucinations during the generation process.
Specifically, 
we first identify the candidates of potential hallucination leveraging the model's logit output values, check their correctness through a validation procedure, mitigate the detected hallucinations, and then continue 
with the generation process.
Through extensive experiments with GPT-3.5 (text-davinci-003) on the `article generation task', we first demonstrate the individual efficacy of our detection and mitigation techniques.
Specifically, the detection technique achieves a recall of $\sim88\%$ and the mitigation technique successfully mitigates $57.6\%$ of the correctly detected hallucinations.
Importantly, our mitigation technique does not introduce new hallucinations even in the case of incorrectly detected hallucinations, i.e., false positives. 
Then, we show that the proposed active detection and mitigation approach successfully reduces the hallucinations of the GPT-3.5 model from $47.5\%$ to $14.5\%$ on average.
We further demonstrate the effectiveness and wide applicability of our approach 
through additional studies including performance on different types of questions (multi-hop and false premise questions) and with another LLM from a different model family (Vicuna). 
In summary, our work contributes to improving the reliability and trustworthiness of large language models, a crucial step en route to enabling their widespread adoption in real-world applications.

\end{abstract}

\section{Introduction}
Recently developed large language models such as GPT-3 \cite{brown2020language}, InstructGPT \cite{ouyang2022training}, PaLM \cite{chowdhery2022palm}, LLaMA \cite{touvron2023llama}, and several others \cite{alpaca, scao2022bloom, 51119, wang-etal-2022-super}
have achieved remarkable performance on a wide range of language understanding tasks.
Furthermore, they have been shown to possess an impressive ability to generate fluent and coherent text. 
Despite all these abilities, \textbf{their tendency to `hallucinate' critically hampers their reliability and limits their widespread adoption in real-world applications}.

Hallucination in the context of language models refers to the generation of text or responses that seem syntactically sound, fluent, and natural but are factually incorrect, nonsensical, or unfaithful to the provided source input \cite{maynez-etal-2020-faithfulness, Holtzman2020The,ji2023survey,koehn-knowles-2017-six}.
These hallucinations can lead to serious consequences such as spreading of misinformation and violation of privacy. 
Thus, \textbf{in this work, we focus on the crucial problem of `\textit{addressing}' large language models' hallucinations.}

\begin{figure}
    \centering
    \includegraphics[width=4.7cm]{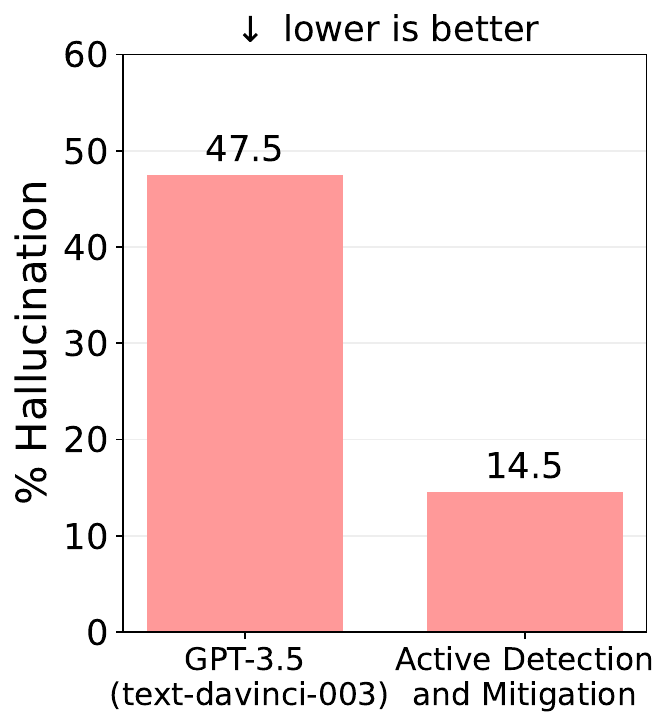}
    \caption{Comparing percentage of hallucinations (on the `article generation task') in the output of GPT-3.5 (text-davinci-003) and our proposed active detection and mitigation approach.}
    \label{fig:approach_mitigation_result}
\end{figure}

\begin{figure*}[!ht]
    \centering 
    \small
    \includegraphics[width=0.97\linewidth]{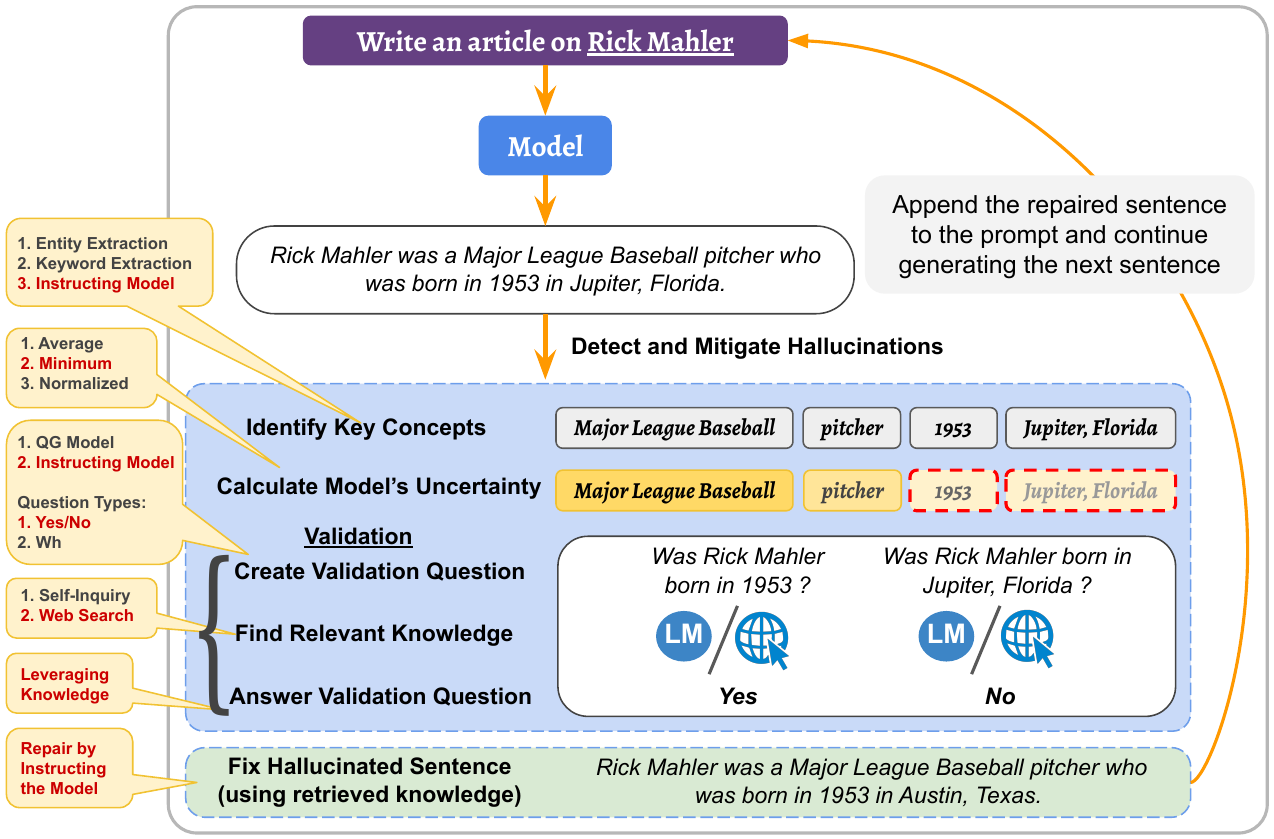}
    \caption{
    Illustration of our proposed approach for addressing LLMs' hallucination problem.
    Given an input, we iteratively generate sentences from the model and actively detect and mitigate hallucinations.
    In the detection stage, we first \textbf{identify the important concepts}, \textbf{calculate model's uncertainty} on them, and then \textbf{validate the correctness} of the uncertain concepts \textbf{by retrieving relevant knowledge}.
    In the mitigation stage, we \textbf{repair the hallucinated sentence} using the retrieved knowledge as evidence. Finally, we append the repaired sentence to the input (and previously generated sentences) and continue generating the next sentence. 
    We show that this procedure not only mitigates current hallucination but also prevents its propagation in the subsequently generated sentences.
    }
    \label{fig:approach}
\end{figure*}

We propose to actively `detect' and `mitigate' hallucinations during the generation process.
{This is crucial as we show that when a generated sentence is hallucinated, the chances of hallucination in the subsequently generated sentences increase.}
Thus, actively detecting and mitigating hallucinations is also important to prevent the propagation of hallucinations in the subsequently generated sentences. We divide our approach into two stages, Detection and Mitigation. 

In the \textbf{hallucination detection} stage, we first identify the candidates of potential hallucination, i.e., the key `concepts' of the generated sentence.
Next, leveraging the logit output values of the model, we calculate model's `uncertainty' on the identified concepts.
We demonstrate that this uncertainty provides a signal for hallucination.
However, we note that this is an additional signal and not a necessary requirement for our approach.
Then, we check the correctness of the 
`uncertain' concepts through a validation procedure where we:
(a) create a query that tests the correctness of the information pertaining to the concept,
(b) retrieve knowledge relevant to the validation question, (c) answer the validation question leveraging the retrieved knowledge, and verify the corresponding information in the generated sentence to detect hallucinations.

This is followed by the \textbf{hallucination mitigation} stage in which we 
`repair' the potentially hallucinated sentence using the retrieved knowledge as evidence.
Figure \ref{fig:approach} illustrates the key steps of our approach.
Furthermore, we conduct a systematic and wide study exploring multiple techniques to achieve the objective of each of the steps.
Importantly, we show that simply instructing the model achieves the corresponding objectives of these steps.

We design an experimental setup where
we prompt the model to write about topics from diverse domains such as sports, politics, music, literature, etc. 
Then, we annotate the correctness of the first five generated sentences for each topic. 
We first demonstrate the individual efficacy of our detection and mitigation techniques.
Specifically, the \textbf{detection technique achieves a recall of $\bm{\sim88\%}$} and the \textbf{mitigation technique successfully mitigates $\bm{57.6\%}$ of the correctly detected hallucinations}.
Importantly, our mitigation technique does not introduce new hallucinations even in the case of incorrectly detected hallucinations, i.e., false positives. 
Then, we show that \textbf{the proposed active detection and mitigation approach successfully reduces the hallucinations of the GPT-3.5 (text-davinci-003) model from $\bm{47.5\%}$ to $\bm{14.5\%}$ on average} (Figure \ref{fig:approach_mitigation_result}).
We further demonstrate the effectiveness and wide applicability of our approach in addressing hallucinations through \textbf{three additional studies}: 
(1) Using another LLM from a different model family (Vicuna-13B),
(2) Adapting the approach to answer multi-hop questions, and 
(3) Assessing it on the `false premise questions'.

\section{Approach}

\subsection{Overview}
\label{method_overview}

We propose to actively detect hallucinations and mitigate them during the generation process.
This is crucial as we show that 
\textbf{a generated sentence is hallucinated more often
when the model has already hallucinated in its previously generated sentences for the input} (Section \ref{hallucination_propagates}).
Similarly, a generated sentence is relatively less often hallucinated when the model has not hallucinated in its previously generated sentences.
Thus, actively detecting hallucinations and mitigating them is also important to prevent the propagation of further hallucinations in subsequently generated sentences. 
To this end, we iteratively generate sentences through the model and actively detect and mitigate hallucinations.
Figure \ref{fig:approach} illustrates the key steps of our approach.

In section \ref{hallucination_detection}, we detail the steps of our hallucination detection approach, i.e., identifying the important `concepts' of the generated sentence, i.e., the candidates of potential hallucination (\ref{keyword_extraction}), calculating model's uncertainty on the concepts using the logit output values (\ref{model_uncertainty}), and checking the correctness by creating validation query (\ref{validation_question}), finding relevant knowledge (\ref{retrieve_knowledge}), and verifying information leveraging the retrieved knowledge (\ref{validation}).
We describe various techniques to achieve the objective of each of these steps and also elaborate on several important points such as 
using a `self-inquiry' method to answer validation questions without using an external knowledge source and trade-off between executing the validation procedure in parallel for all the concepts and in sequential order based on their `uncertainty'.
For each step, we also \textbf{indicate the most preferred technique with (*)} and provide our justification.

In section \ref{hallucination_mitigation}, we detail our hallucination mitigation approach.
Specifically, we `repair' the hallucinated sentence by removing or substituting the hallucinated information leveraging the retrieved knowledge as evidence, and can also utilize the retrieved knowledge as context (prepended to the input) to generate the next sentence.

\subsection{Hallucination Detection}
\label{hallucination_detection}

\subsubsection{Identify Key Concepts}
\label{keyword_extraction}

In the first step, we identify the important concepts from the generated sentence. 
We identify these concepts because validating the correctness of the entire sentence at once is infeasible; this is because a sentence may contain a number of different facets all of which can not be validated at once.
On the other hand, individually validating the correctness corresponding to the concepts provides opportunities for accurately detecting hallucinations.
Thus, the objective of this step is to identify the candidates of potential hallucination.
We note that a concept or keyphrase is essentially a span of text consisting of one or more words.
We study the following techniques to identify the concepts:

\paragraph{Entity Extraction:}
Entities are usually an important part of a sentence, thus, we use an off-the-shelf entity extraction model 
to identify the concepts.
A limitation of this method is that a concept need not necessarily be an entity and can be a non-entity span also. 
We address this limitation with a keyword extraction model.

\paragraph{Keyword Extraction:}
To also identify the non-entity concepts, we explore an off-the-shelf keyword extraction model\footnote{https://huggingface.co/ml6team/keyphrase-extraction-kbir-kpcrowd}. 
This model uses Keyphrase Boundary Infilling with Replacement (KBIR) as its base model and fine-tunes it on the KPCrowd dataset \cite{kulkarni2021learning}.

\paragraph{*Instructing the Model*:}
Since state-of-the-art language models perform remarkably well on a wide range of tasks, in this technique, we directly instruct the model to identify the important concepts from the generated sentence. 
An important characteristic of this technique is that it doesn't require calling a task-specific tool (entity or keyword extraction model) for this task. 

Table \ref{tab:keyword_extraction_examples} (in Appendix \ref{examples_keywords}) illustrates examples of concepts identified using the three techniques. 
It shows that the entity extraction model misses many important concepts while the keyword extraction model identifies a lot of insignificant concepts also.
In contrast, instruction technique successfully identifies all the important concepts.
Moreover, it doesn't require calling a task-specific tool. 
Thus, we represent this technique with (*), our preferred technique for this step.

\subsubsection{Calculate Model's Uncertainty}
\label{model_uncertainty}

GPT-3 \cite{brown2020language} and several other publicly available models also provide logit output values in their prediction response. 
Thus, we study if these logit output values can be utilized to 
detect hallucinations.
However, we note that this is an additional source of information and not a necessary requirement for our hallucination detection method as some models that are available only via API calls do not provide these logit output values.

Recall that a concept can consist of more than one token also (note that the model provides logit output values at the level of tokens); thus, we study three different techniques for calculating a \textbf{probability score} for a concept.
Consider a concept consisting of $n$ tokens and having the maximum softmax probabilities as $p_1, p_2, p_3, ..., p_n$ for the $n$ token positions respectively.
We obtain these probabilities by applying the softmax function over the logit values for each token position.
We study the following techniques:

\paragraph{Average of Token Probabilities}:
In this technique, we simply take the average of the probabilities of the tokens corresponding to the concept:
\[
     \text{score} = \text{AVG} (p_1, p_2, ..., p_n)
\]

\paragraph{Normalized Product of Token Probabilities}:
Here, we take a normalized product of the probabilities of the tokens:
\[
     \text{score} = (p_1 \times p_2 \times ... \times p_n)^{1/n}
\]

\paragraph{*Minimum of Token Probabilities*}:
Here, we take the minimum of probabilities as the score.
\[
     \text{score} = \text{MIN} (p_1, p_2, ..., p_n)
\]

This is our preferred technique for this step as the other techniques average out the effect of model's uncertainty on the tokens while low probability in even one token of the concept provides a strong evidence of the model being uncertain.
For example, if the model is uncertain on the name of the USA president then its uncertainty on the first token (`Joe') would be high but on the next token (`Biden') would be very low as the token `Joe' is frequently followed by the token `Biden'. 
Thus, averaging or normalizing the probabilities will have a limited capability to capture this signal.

Through our experiments (Section \ref{uncertain_keyphrases_hallucinated}), we show that this score (especially `MIN') indeed provides a signal for hallucination, i.e., the more uncertain a model is on a concept (low probability score), the more likely it is to be hallucinating about that concept.
However, we note that this score is just a signal for hallucination and in no way provides a guarantee for presence of hallucinations.
We utilize this signal and check for hallucinations with respect to the uncertain concepts using our validation procedure (\ref{validation_question}-\ref{validation}).

\paragraph{In the absence of logit output values:}
For models that do not provide the logit output values, all or some heuristically selected concepts (depending on the computational and latency budget of the system) can be passed to the validation stage for detecting hallucinations.

\subsubsection{Create Validation Question}
\label{validation_question}

We start the validation procedure for a concept by creating a question that tests the correctness of the information (in the generated sentence) pertaining to the concept.
We create \textbf{Yes/No Questions}, i.e., questions for which the answer is either a `Yes' or a `No'. 
Table \ref{tab:validation_question_examples} shows examples of validation questions.
For creating these questions, we explore the following two techniques:

\paragraph{Question Generation Tool: }
Here, we use an off-the-shelf answer-aware question generation model. 

\paragraph{*Instructing the Model*:}
Here, we directly instruct the model to create a validation question checking the correctness of the information about the selected concept.
For the same reason as in the concept identification step, this is our preferred technique as it does not require calling a task-specific tool.

We note that instead of Yes/No questions, \textbf{Wh-questions} can also be used for validation. 
We prefer Yes/No questions as it is relatively easier to check the answer for these questions.
We explore Wh-questions for a case study for answering multi-hop questions (Section \ref{case_study_multihop_questions}).

\subsubsection{Find Relevant Knowledge}
\label{retrieve_knowledge}

\paragraph{*Web Search*:}
In order to answer the validation question, we retrieve knowledge relevant to it which serves as additional context.
For generality and wide coverage, we use web search (via Bing search API) for retrieving this knowledge. 
However, we note that any other search API or knowledge corpus can also be utilized for this purpose.

\paragraph{Self-Inquiry:}
We also explore a self-inquiry technique where we directly prompt the model to answer the validation question.
In this technique, the model relies on its parametric knowledge to answer the validation question. 
This technique has several drawbacks as compared to web search such as lack of a reliable strategy to extract the parametric knowledge from the model and staleness of the parametric knowledge.


Note that the proposed knowledge retrieval step in our approach has several benefits, such as (a) it does not retrieve knowledge when it is not required, i.e., when the model is already sufficiently confident (since we show it is less likely to hallucinate in such scenarios),
(b) it individually retrieves knowledge pertinent to the concept(s) on which the calculated probability score is low thus providing it sufficient and relevant context for accurate validation / mitigation.

\subsubsection{Answer Validation Question}
\label{validation}

In this step, we prompt the model to answer the validation question (leveraging the retrieved knowledge as context) and verify its response.
If the validation procedure succeeds for all the uncertain concepts then we continue generating the next sentence; otherwise, we interrupt the generation process, mitigate the potential hallucination in the sentence, and then continue generation.

\paragraph{Order of Validation of Concepts:}
Validation of different concepts can be done in a sequence (in ascending order of their calculated probability score) or in parallel.
However, running this in parallel would require starting multiple threads which may not be supported by all machines. 
Thus, in this work, we study only the sequential validation strategy but note that it can be made more efficient by running it in parallel.
We regard this sequential validation as a greedy exiting strategy as we proceed to the mitigation stage on detection of the first potential hallucination.

\subsection{Hallucination Mitigation}
\label{hallucination_mitigation}

For mitigating the hallucination in the generated sentence, we instruct the model to repair the generated sentence by either removing or substituting the hallucinated information using the retrieved knowledge as evidence.
Table \ref{tab:prompts} shows the instructional prompts for different steps of our approach.

\textbf{Note:} We note that the result of the validation procedure is contingent on the retrieved knowledge and the model's ability to leverage that knowledge in answering the validation question.
Thus, a case is plausible in which the validation procedure reports hallucination even though the sentence is actually not hallucinated.
However, in Section \ref{hallucination_detection_perf}, we show that our approach performs fairly well on this task. 
Moreover, it achieves a very high recall demonstrating its efficacy at detecting hallucinations.
Moreover, in Section \ref{hallucination_mitigation_perf}, we show that our mitigation approach does not introduce new hallucinations even in the case of incorrectly detected hallucinations, i.e., false positives.

\subsection{Design Decisions}

\paragraph{Why the task of addressing hallucinations is broken down into several steps?}

We note that dealing with the hallucination problem is a complex task and prior work has shown that breaking down a complex task into simpler sub-tasks helps the model in solving the task \cite{wei2022chain,zhou2023leasttomost,khot2023decomposed}. 
Thus, we break down this task into individual sub-tasks which are considerably easier for the model.
For the same reason, we also break down the validation procedure into several steps.

\paragraph{Why validation is done using the web search?}

Our preferred technique for retrieving knowledge is web search because the web is more likely to contain the updated knowledge in comparison to a knowledge corpus whose information can become stale, outdated, and obsolete.

\paragraph{Why ``active'' detection \& mitigation and not ``post-hoc'' after complete response generation?}

We note that our detection and mitigation techniques can also be applied in a ``posthoc'' manner after complete response generation. However, it has several limitations which are addressed by our ``active'' approach.
The ``active'' approach prevents the propagation of hallucinations in the subsequently generated sentences, i.e., if hallucination is detected in the initially generated sentences then it would be mitigated and course correction would be done for the subsequently generated sentences. However, the ``post-hoc'' approach does not provide such an opportunity of course correction.
In other words, in the ``active'' approach, the model sees the mitigated / corrected sentences while generating the subsequent sentences; thus, its output will be more correct, coherent, and fluent. In contrast, in the ``posthoc'' approach, the generated sentences are based on the initially generated previous sentences and thus the mitigated sentence will not be able to influence the generation of subsequent sentences; thus, the output would not be as coherent and fluent as the active approach.

\paragraph{Impact on Inference Cost: }

Our approach results in improvements in the form of reduced hallucinations and thus makes the model more reliable; however, it comes at the expense of increased inference cost. 
However, we believe that at current time, to enable the widespread adoption of LLMs, it is more important to address their reliability and trustworthiness concerns because computational advancements are ongoing at a rapid pace. Moreover, even larger models with multi-fold times more parameters such as PaLM (540B) \cite{chowdhery2022palm}, Gopher (280B) \cite{rae2021scaling}, and MT-NLG (530B) \cite{smith2022using} are also being developed which have even higher inference cost showcasing a larger focus of the community on developing better performing systems.
However, we note that our approach can be made more efficient by various techniques discussed before such as validating concepts in parallel and executing these intermediate steps using a smaller low-cost model.

\section{Experiments and Results}
\label{sec_experiments}

In this section, we first demonstrate the two findings that motivate our approach (\ref{hallucination_propagates} and \ref{uncertain_keyphrases_hallucinated}).
Then, we show the individual efficacy of our hallucination detection and mitigation techniques in \ref{hallucination_detection_perf} and \ref{hallucination_mitigation_perf}, respectively.
Finally, in \ref{approach_performance}, we show the effectiveness of the proposed active detection and mitigation approach in addressing hallucinations.

\paragraph{Data and Annotation:}

\begin{figure}
    \centering
    \includegraphics[width=7.5cm]{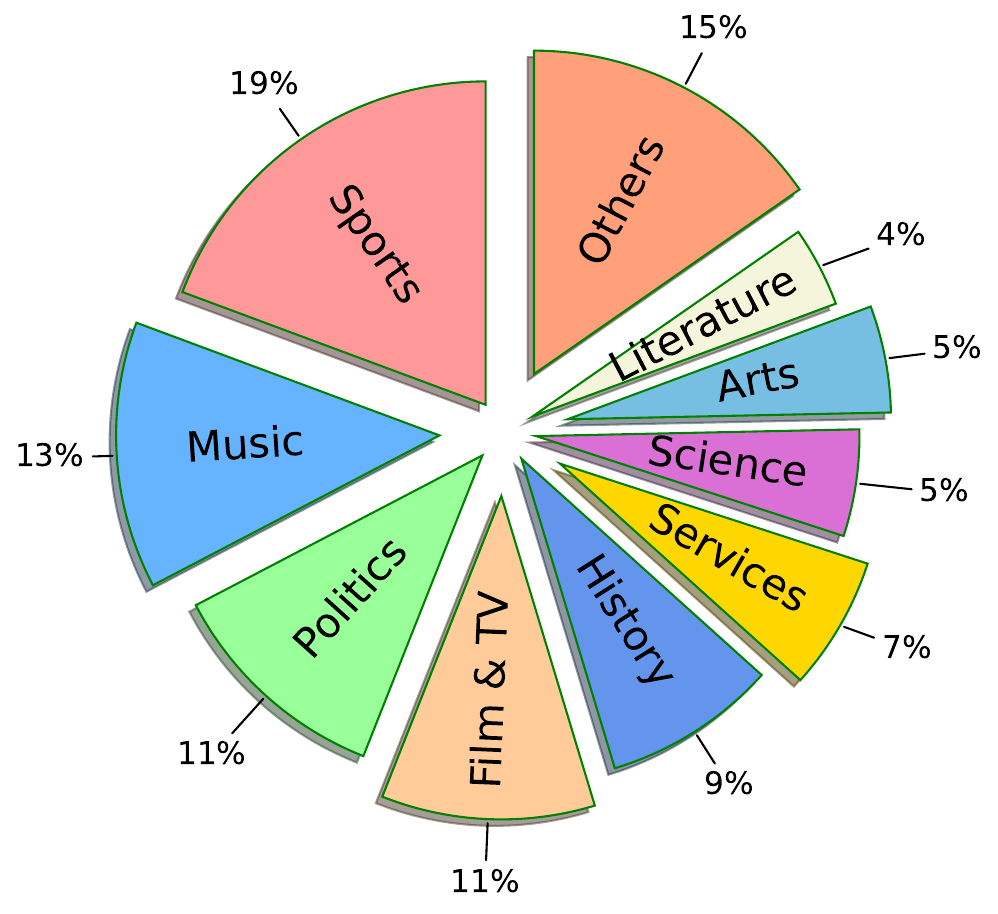}
    \caption{Distribution of instances across different domains in our topic set.}
    \label{fig:data_type_1_categories}
\end{figure}

In our experimental setup, we prompt the large language model (GPT-3.5: text-davinci-003) to write about various topics.
Specifically, we use a total of $150$ topics from diverse domains.
Figure \ref{fig:data_type_1_categories} shows the distribution of different domains in our topic set.
In each domain, we include different kinds of topics; for instance, Sports domain consists of sports persons, administrators, teams, and games, Music consists of musicians, songs, music labels, and bands, Politics includes politicians, political parties, and elections, Film \& TV includes actors, TV personalities, shows, and movies, History includes historians and events, etc.
For selecting the names of people, we use randomly sampled names from the top 20\% of longest articles in WikiBio dataset \cite{lebret-etal-2016-neural} as done in \cite{manakul2023selfcheckgpt}.
Similarly, for the other topics, we randomly sample from the longest Wikipedia articles.
This is done to ensure that no obscure or ambiguous concept is selected.

\begin{figure*}
    \centering
    \includegraphics[width=14.5cm]{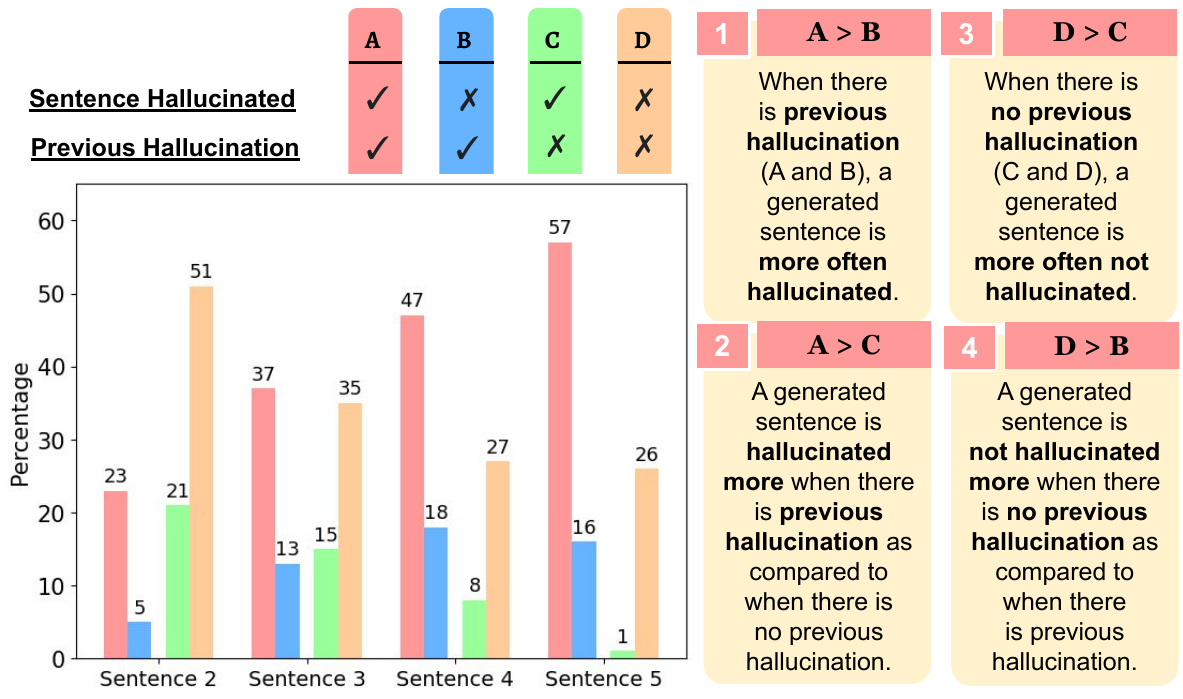}
    \caption{Demonstrating relationship between `hallucination in a generated sentence' and `hallucination in the previously generated sentences'. Bars A, B, C, and D correspond to the four possibilities of the relationship between the two binary variables.
    On the right, we mention our four inferences from the figure.}
    \label{fig:hallucination_causes_hallucination}
\end{figure*}
Equipped with the list of topics, we give the following input prompt to the model:
\texttt{``Write an article about <topic>''} for each topic.
Following this, we (the authors) manually annotate the correctness of the first five sentences generated by the model for each topic. 
For annotating the correctness, we look at search results from the web to find the relevant knowledge that either supports or contradicts the information present in the generated sentence. 
In some cases, multiple web searches were required to check the correctness of different facets of a sentence. 
Furthermore, in a small number of cases, we could not find information supporting or contradicting the information in the generated sentence, we mark it as a case of extrinsic hallucination. 
We opt for this expert annotation strategy because despite our annotation task being a simple binary classification task, it requires considerable effort in checking the correctness of a given sentence which can not reliably be collected via crowdsourcing.
In addition to this sentence-level annotation, we also annotate correctness at the concept-level that we will detail in \ref{uncertain_keyphrases_hallucinated}.
We release both sentence-level and concept-level hallucination annotations that will also facilitate a systematic future research in this direction.

\begin{figure*}[t]
\centering
    \begin{subfigure}{.45\textwidth}
        \includegraphics[width=\linewidth]{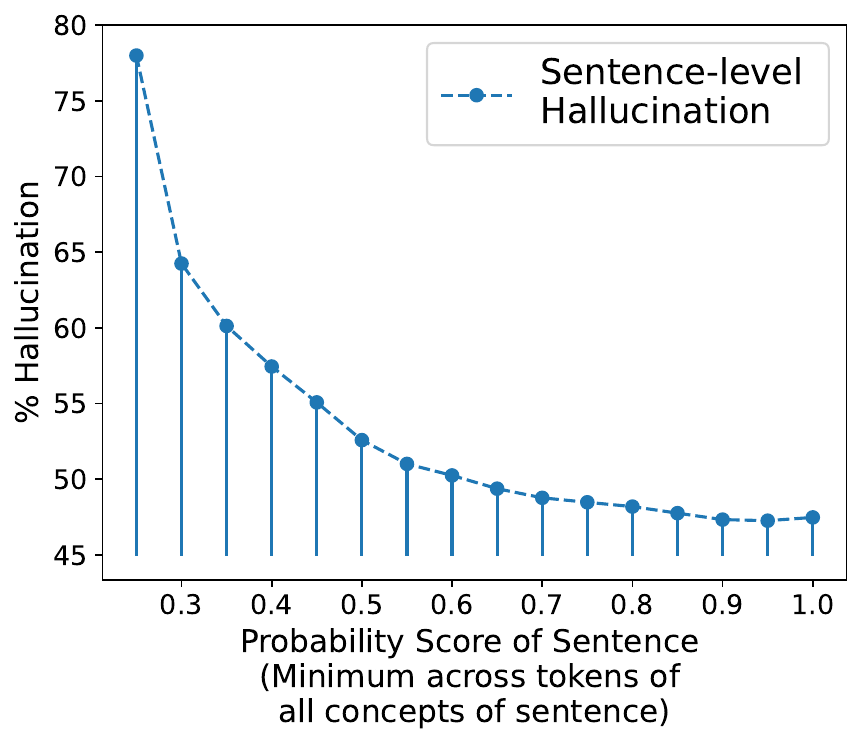}
    \end{subfigure}
    \begin{subfigure}{.45\textwidth}
        \includegraphics[width=\linewidth]{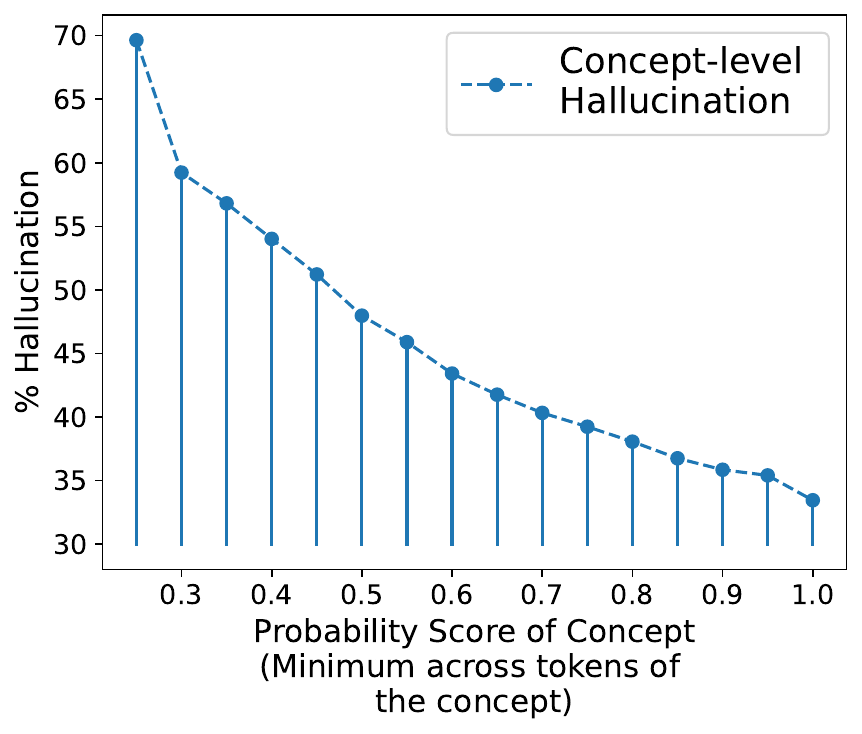}
    \end{subfigure}
    
    \caption{Trend of hallucination with the calculated probability score (Minimum technique) at both sentence and concept level. \textbf{As the probability increases, the model's tendency to hallucinate decreases}.
    }
    \label{fig:comparing_percentage_hallucination_with_keyphrase_prob}    
\end{figure*}

\subsection{Motivating Findings}
\label{motivating_findings_sec}

\subsubsection{Hallucination Causes Further Hallucination}
\label{hallucination_propagates}

Recall that we consider the first five sentences generated by the model for each topic and annotate their correctness.
Since the sentences are sequentially generated, we investigate the 
relationship between `hallucination in a generated sentence' and `hallucination in the previously generated sentences' for an input. 
Since there are two binary variables, there exist four possibilities in this relationship, i.e., 
a sentence is hallucinated and there was hallucination in the previously generated sentences \textbf{(A)}, the sentence is not hallucinated and there was hallucination in the previously generated sentences \textbf{(B)}, the sentence is hallucinated and there was no hallucination in the previously generated sentences \textbf{(C)}, the sentence is not hallucinated and there was no hallucination in the previously generated sentences \textbf{(D)}. 
For illustration, consider a sample case for sentence 3, the two binary variables are whether sentence 3 is hallucinated and whether there was hallucination in the previously generated sentences (i.e. in sentence 1 OR sentence 2).
Figure \ref{fig:hallucination_causes_hallucination} demonstrates this relationship for sentences 2, 3, 4 and 5 aggregated over all the topics in our data.
We do not show this for sentence 1 as there is no previously generated sentence for it.

From this figure, we draw the following inferences:

(a) \textbf{A > B}: Cases A and B correspond to the scenario when there is hallucination in the previously generated sentences. It can be observed that A is considerably greater than B which implies that \textit{when there is hallucination in the previously generated sentences, a sentence is hallucinated more often}. 
Moreover, the gap keeps increasing as the sentence number increases.

(b) \textbf{A > C}: Cases A and C correspond to the scenario when a generated sentence is hallucinated. It can be observed that A is greater than C which implies that \textit{a generated sentence is hallucinated more when there is hallucination in the previously generated sentences as compared to when there is no previous hallucination}.

(c) \textbf{D > C}: Cases C and D correspond to the scenario when there is no hallucination in the previously generated sentences. Here, D is greater than C which implies that \textit{when there is no hallucination in the previously generated sentences, a generated sentence is more often not hallucinated}.

(d) \textbf{D > B}: Cases B and D correspond to the scenario when a generated sentence is not hallucinated. D is greater than B which implies that \textit{a generated sentence is not hallucinated more when there is no previous hallucination as compared to when there is previous hallucination}.

This shows that hallucination in a sentence often results in further hallucinations in the subsequently generated sentences and thus \textbf{actively detecting and mitigating hallucinations can not only fix the current hallucination but can also prevent its propagation in the subsequently generated sentences}.

Next, we demonstrate the utility of logit output values in detecting hallucinations.

\subsubsection{Logit Output Values Provide a Signal for Hallucination}
\label{uncertain_keyphrases_hallucinated}

In this subsection, we first show the trend of hallucination with the probability score.
Note that this score is calculated using the logit output values.
Then, we demonstrate the benefit of identifying concepts from the generated sentence in detecting hallucinations.
Finally, we compare the efficacy of different probability calculation techniques in detecting hallucinations.

\paragraph{Hallucination vs Probability Score:}
In order to study the relationship between logit output values and hallucination, we annotate correctness at concept-level also (in addition to sentence-level annotations described earlier).
Specifically, for each identified concept, we mark whether the information about it in the generated sentence is hallucinated or not. 
This can be different from sentence-level annotation as it focuses only on the correctness of the information about the concept in the sentence.
Table \ref{tab:sentence_annotation_example} shows examples of both sentence-level and concept-level annotations.

Figure \ref{fig:comparing_percentage_hallucination_with_keyphrase_prob} shows the trend of hallucination with our calculated probability scores at both sentence and concept levels.
For a sentence, we use the minimum across tokens of all its identified concepts as the probability score and for a concept, we use the minimum across all its tokens as the probability score.
It can be observed that \textbf{as the probability score increases (or uncertainty decreases), tendency to hallucinate decreases.}
This shows that these probability values can be utilized as a signal for hallucination, i.e., the low probability concepts in a generated sentence can be considered as candidates of potential hallucination and their correctness in the generated sentence can be validated for detecting hallucinations. 
On average, we observe an absolute difference of $\sim0.15$ between the probabilities of concepts when the model is hallucinating vs when it is not hallucinating.

\begin{figure}
    \centering
    \includegraphics[width=7.cm]{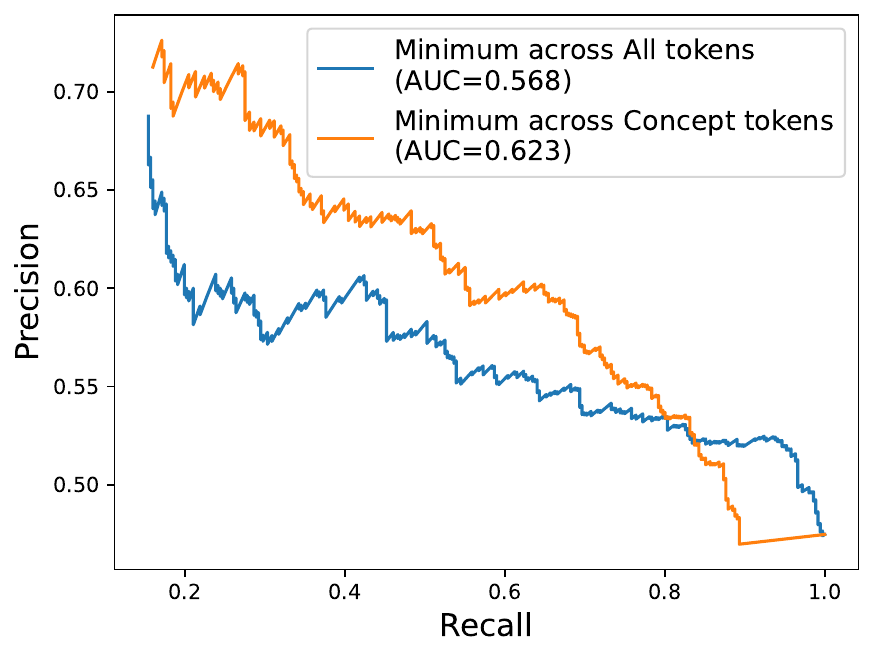}
    \caption{
    Demonstrating the benefit of identifying concepts from a sentence for detecting hallucinations.
    The figure shows precision-recall curves for the sentence level hallucination detection task corresponding to two methods that use the probabilities calculated from the logit output values. 
    The blue curve corresponds to the technique in which we use the minimum probability across \textbf{all tokens} of the sentence and the orange curve is for the technique in which we use the minimum over \textbf{only the tokens of the identified concepts}.
    }
    \label{fig:comparing_detection_performance_of_all_tokens_vs_keyphrase_tokens}
\end{figure}

\paragraph{Benefit of Identifying Concepts from a Sentence:}
Now, we demonstrate the benefit of identifying concepts from a sentence and leveraging the logit output values corresponding to their tokens for detecting hallucinations.
To this end, we plot precision-recall curves for the hallucination detection task corresponding to two methods that use the probabilities calculated from the logit output values. 
The blue curve corresponds to the technique in which we use the minimum probability across \textbf{all tokens} of the sentence and the orange curve is for the technique in which we use the minimum over \textbf{only the tokens of the identified concepts}.
Figure \ref{fig:comparing_detection_performance_of_all_tokens_vs_keyphrase_tokens} shows the two curves.
The orange curve achieves higher area under the precision-recall curve implying that utilizing \textbf{the probabilities of the concept tokens provides a stronger signal for hallucination} as compared to the probabilities corresponding to all the tokens.

\begin{figure}
    \centering
    \includegraphics[width=7.cm]{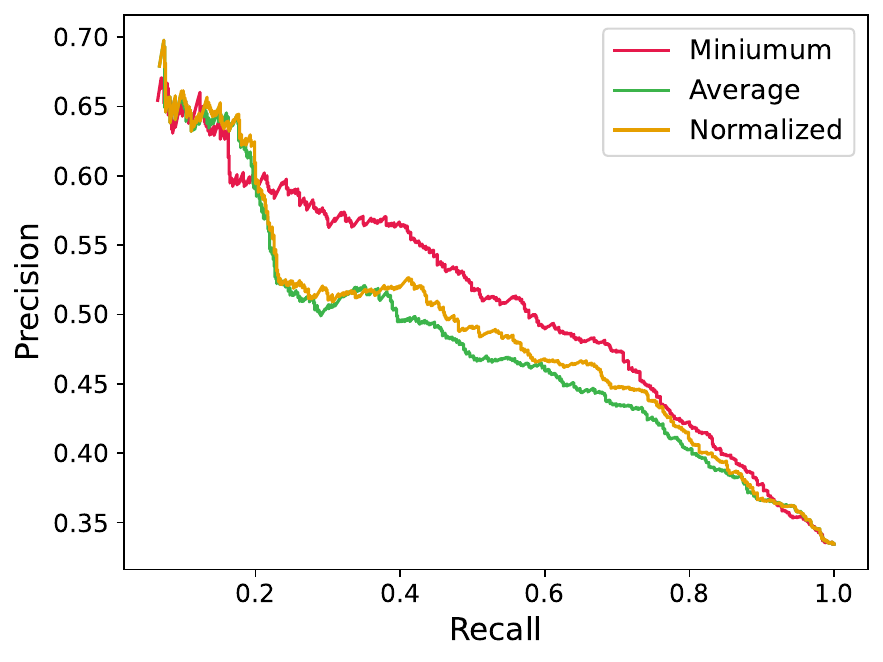}
    \caption{
    Precision-Recall curves for the hallucination detection task (at {concept-level}) using the three probability calculation techniques.
    \textbf{The `Minimum' technique achieves the highest AUC.}
    }
    \label{fig:keyphrase_level_min_avg_normalized_comparison}
\end{figure}

\paragraph{Comparing Probability Calculation Techniques:}
Figure \ref{fig:keyphrase_level_min_avg_normalized_comparison} shows the Precision-Recall curves for the hallucination detection task (at concept-level) using the three probability calculation techniques, i.e., Minimum, Average, and Normalized (described in \ref{model_uncertainty}). 
\textbf{The `Minimum' technique achieves the highest area under the curve and hence is better at the hallucination detection task.}

\subsection{Hallucination Detection Performance}
\label{hallucination_detection_perf}

In this subsection, we demonstrate the hallucination detection performance of various techniques at both sentence and concept-levels.

\paragraph{Self-Inquiry vs Web Search:}
Table \ref{tab:sentence_level_detection_accuracy_self_inquiery_vs_web_search} and \ref{tab:keyword_level_detection_accuracy_self_inquiery_vs_web_search}
show the hallucination detection performance of the self-inquiry and web search techniques at sentence-level and concept-level, respectively. 
For sentence-level results, we predict the sentence to be hallucinated if the validation procedure fails on any identified concept.  
Note that in these results, we do not leverage the uncertainty score to select concepts for validation, instead we validate all the identified concepts.
We study the relationship of recall with probability thresholds in Figure \ref{fig:recall_vs_prob_self_inquiry_and_web_search_comparison} (in Appendix).
From the tables, it can be observed that the web-search technique achieve considerably high recall in detecting hallucinations.

Here, we emphasize on the high `recall' of web-search technique as we show that our mitigation approach does not introduce any new hallucinations even in the case of incorrectly detected hallucinations, i.e., false positives (\ref{hallucination_mitigation_perf}). 
Figure \ref{fig:recall_vs_prob_self_inquiry_and_web_search_comparison} shows the recall of hallucination detection vs Probability threshold plot for Self Inquiry and web search techniques at both sentence-level and concept-level.
\textbf{Web-search is consistently and considerably better than self-inquiry.}

\begin{table}[]
\begin{subtable}[t]{0.48\textwidth}
  \caption{\textbf{Sentence level}}

\centering
\small
\resizebox{0.98\linewidth}{!}
{
\begin{tabular}{c|c|cccc}
\toprule
\multirow{2}{45pt}{\textbf{Technique}}
& \multirow{2}{45pt}{\textbf{Accuracy}} & \multicolumn{2}{c}{\textbf{Hallucinated}} & \multicolumn{2}{c}{\textbf{Not Hallucinated}} \\ 

& & \textbf{Prec.} & \textbf{Rec.} & \textbf{Prec.} & \textbf{Rec.} \\
\midrule


\textbf{Self-Inquiry} & 0.62 & 59.89 & 63.76 & 65.23 & 61.42 \\


\textbf{Web-Search} & {0.681} & {61.82} & \textbf{85.96} & {80.39} & 52.03 \\

\bottomrule
\end{tabular}
}

  \label{tab:sentence_level_detection_accuracy_self_inquiery_vs_web_search}
\end{subtable}

\begin{subtable}[t]{0.48\textwidth}
\caption{\textbf{Concept level}}
  
\centering
\small
\resizebox{0.98\linewidth}{!}
{
\begin{tabular}{c|c|cccc}
\toprule
\multirow{2}{45pt}{\textbf{Technique}}
& \multirow{2}{45pt}{\textbf{Accuracy}} & \multicolumn{2}{c}{\textbf{Hallucinated}} & \multicolumn{2}{c}{\textbf{Not Hallucinated}} \\ 

& & \textbf{Prec.} & \textbf{Rec.} & \textbf{Prec.} & \textbf{Rec.} \\
\midrule


\textbf{Self-Inquiry} & 0.65 & 47.96 & 45.85 & 73.37 & 74.98 \\


\textbf{Web-Search} & {0.75} & 58.17 & \textbf{87.68} & 91.69 & 68.30 \\

\bottomrule
\end{tabular}
}
  \label{tab:keyword_level_detection_accuracy_self_inquiery_vs_web_search}

\end{subtable}

\caption{Hallucination detection performance of self-inquiry and web-search techniques at both sentence and concept levels. It also shows separate precision and recall on both hallucinated and non-hallucinated instances.}
\label{tab:detection_accuracy_self_inquiery_vs_web_search}
\end{table}

\subsection{Hallucination Mitigation Performance}
\label{hallucination_mitigation_perf}





\begin{table}[]
\small
\begin{tabular}{cc|c}
\toprule
\textbf{Before}  & \textbf{After}         & \textbf{Percentage} \\
\midrule
Hallucinated     & Not Hallucinated       & \colorbox{green}{40.81\%}               \\
Hallucinated     & Hallucinated     & 30.04\%               \\ 
Not Hallucinated & Not Hallucinated & \colorbox{green}{28.26\%}               \\
Not Hallucinated & Hallucinated           & \colorbox{coralred}{0.89\%}    \\
\bottomrule

\end{tabular}
\caption{Result on modifying the reported hallucinations. Our approach successfully mitigates hallucinations on $57.6\%$ of the correctly detected hallucinations while deteriorating a minimal $3.06\%$ of the incorrectly detected hallucinations i.e. false positives.
}
\label{tab:mitigation_perf}
\end{table}

On sentences where our validation procedure (using Web search) reports hallucinations, we apply our mitigation technique.
We note that a sentence which is reported as hallucination can either be actually hallucinated or not hallucinated, i.e., it could also be a false positive.
Table \ref{tab:mitigation_perf} shows the result of our method.
It successfully mitigates the hallucination on $57.6\%$ of the correctly detected hallucinations (True Positives); we refer to this metric as `success'.
Furthermore, it achieves this at minimal `deterioration' ($3.06\%$), i.e., it incorrectly converts a minimal $3.06\%$ of the non-hallucinated instances to hallucinated.
Table \ref{tab:mitigation_examples} (in Appendix) shows examples where our mitigation technique successfully mitigates the hallucinations.

\paragraph{Analyzing Failures in Mitigating Hallucinations:}
Table \ref{tab:incorrect_mitigation_examples} (in Appendix) shows examples where our mitigation technique fails to mitigate the hallucinations. 
We observe that in many of the failure cases, our technique fixes some hallucinated content of the sentences but fails to fix ALL the hallucinated content from them.
For instance, example 1 and 2 in Table \ref{tab:incorrect_mitigation_examples} correspond to type of failure.

Furthermore, in some of the failure cases, our technique results in a sentence which is no longer hallucinated but it not completely related to the topic.
For instance, the fourth example in Table \ref{tab:incorrect_mitigation_examples} about the topic `Harry S. Kennedy'; the model generates ``\textit{Harry S. Kennedy was an American politician who served as the 35th President of the United States from 1961 to 1963.}'' which is wrong and our mitigation technique modifies it to ``\textit{John F. Kennedy was an American politician who served as the 35th President of the United States from 1961 to 1963.}'' which is factually correct but not related to the topic `Harry S. Kennedy'. This happens because the output of the mitigation step is contingent on the information in the retrieved knowledge.

\begin{figure}
    \centering
    \includegraphics[width=7.3cm]{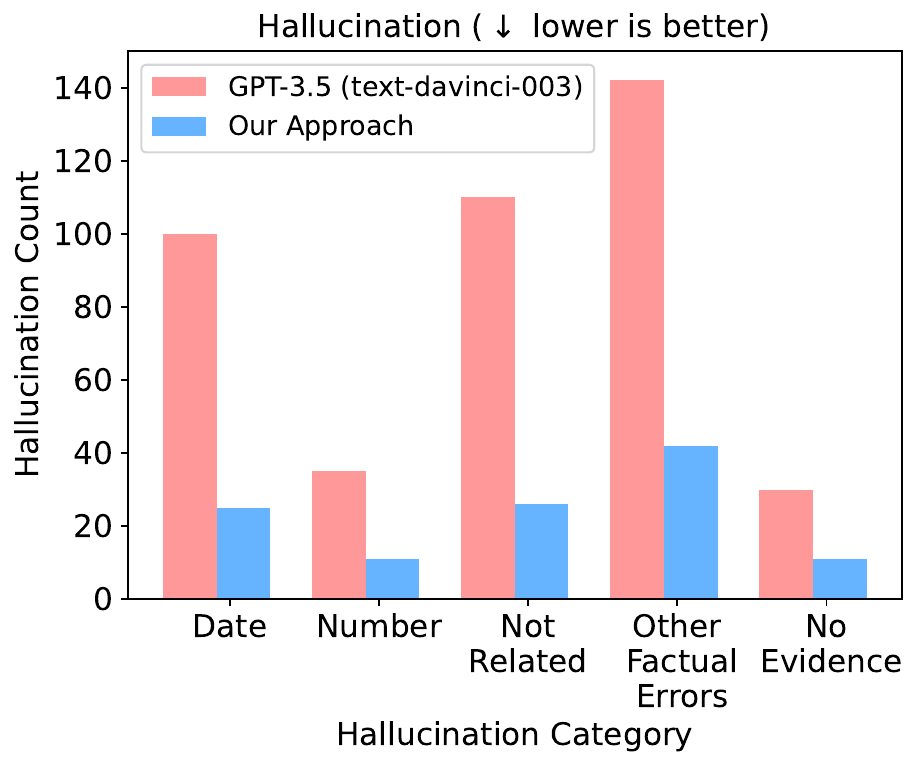}
    \caption{Comparing number of hallucinations across different categories for sentences generated using GPT-3.5 and our active detection and mitigation approach.}
    \label{fig:hallucination_categories_pie_chart}
\end{figure}

\subsection{Active Detection and Mitigation}
\label{approach_performance}

The two findings in Section \ref{motivating_findings_sec} motivate our approach of addressing hallucinations in which we actively detect hallucinations leveraging the logit output values and mitigate them during the generation process to prevent their propagation. 
Specifically, using the calculated probability scores, we identify the uncertain concepts and check their correctness using our validation procedure.
We generate one sentence at a time and when our detection method reports hallucination, we fix it using our mitigation approach and continue generating the next sentence.
We demonstrated separate detection and mitigation efficacy in \ref{hallucination_detection_perf} and \ref{hallucination_mitigation_perf}, respectively.
Figure \ref{fig:approach_mitigation_result} compares the percentage of hallucination in the output of GPT-3.5 model and our active detection and mitigation approach.
Our approach reduces the percentage of hallucinations from $47.4\%$ to $14.53\%$.
In Figure \ref{fig:hallucination_categories_pie_chart}, we demonstrate this comparison for different categories of hallucination.
It shows that our approach reduces hallucinations for all categories.

To further demonstrate the effectiveness and wide applicability of our approach, we present three interesting additional studies. 
In the first study (Section \ref{case_study_other_llms}), we experiment with \textbf{another large language model, Vicuna-13B} \cite{vicuna2023} and show that our approach performs well with this model also and {considerably reduces the hallucinations it its output} (Figure \ref{fig:approach_mitigation_result_vicuna_13B}). 
We select this model since it is widely popular and publicly available to use.
In the second and third studies, we adapt our approach to two different types of questions and show its effectiveness.
Specifically, in Section \ref{case_study_multihop_questions}, we adapt it to answer \textbf{multi-hop questions}
and in Section \ref{case_study_false_premise_questions}, we show experiment with the \textbf{false premise questions}.

\section{Additional Experiments}

\subsection{Efficacy with LLM from Another Family}
\label{case_study_other_llms}

\begin{figure}
    \centering
    \includegraphics[width=4.7cm]{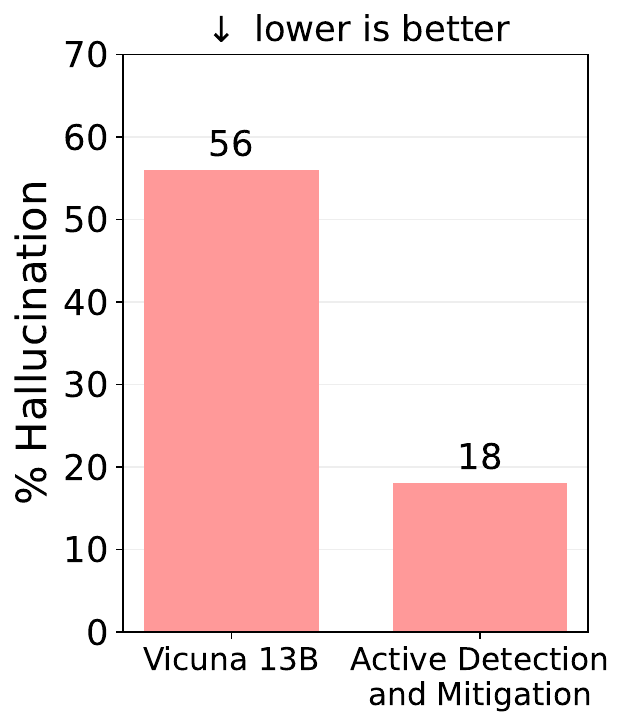}
    \caption{Comparing percentage of hallucinations (on the `article generation task') in the output of \textbf{Vicuna-13B} model and our proposed approach.}
    \label{fig:approach_mitigation_result_vicuna_13B}
\end{figure}

Figure \ref{fig:approach_mitigation_result_vicuna_13B} compares the percentage of hallucinations (on the `article generation task') in the output of Vicuna-13B and our proposed active detection and mitigation approach.
It shows that our approach considerably reduces the hallucinations similar to the case with GPT-3.5 model. 
This study is conducted on $10$ randomly sampled topics (i.e, $50$ generated sentences) from the topic set described in Section \ref{sec_experiments}.
We note that similar to the setup with the GPT-3.5 model where we used instructional prompts with GPT-3.5 for all the steps of the approach (such as identifying key concepts and creating validation questions), here, in this setup, we use the Vicuna-13B model itself for all those steps.
This result demonstrates the generality and applicability of our approach for other models also.

\subsection{Multi-hop Questions}
\label{case_study_multihop_questions}

\begin{table*}[]
\centering
\small
{
\begin{tabular}
{@{}p{0.99\linewidth}}
\toprule


    \textbf{Question}: \textit{Kent Dairy Round Barn is near the city in what Montana county?} \\
    
    {\textbf{Response}: The Kent Dairy Round Barn is located near Red Lodge, Montana. Red Lodge is located in Carbon County, Montana. So, the answer is \colorbox{green}{Carbon County}.} \\ \\
    
    \textbf{Question}: \textit{What political party was the 6th governor of Hawaii that passed bills to help with global warming in Hawaii?}  \\
    
    {\textbf{Response}: The 6th governor of Hawaii was Linda Lingle. Linda Lingle was a member of the Republican Party and passed bills to help with global warming in Hawaii. So, the answer is \colorbox{green}{Republican Party}.} \\ \\ 
    
    \textbf{Question}: \textit{In what city is the mortgage lending company, founded by Dan Gilbert, headquartered?} \\
    
    {\textbf{Response}: The mortgage lending company founded by Dan Gilbert is Quicken Loans. Quicken Loans is headquartered in Detroit, Michigan. So, the answer is \colorbox{green}{Detroit, Michigan}.} \\ \\
    
    \textbf{Question}: \textit{Actor David Lee Stenstrom played the character Waldo the inventor in a TV show that ran on Nickelodeon during what yeras?} \\
    
    {\textbf{Response}: The TV show was Out of Control (1984). David Lee Stenstrom played the character Waldo the inventor in the show. The show ran on Nickelodeon from 1984 to 1985. So, the answer is \colorbox{green}{1984 to 1985}.} \\  \\

    \textbf{Question}: \textit{What music school did the singer of The Search for Everything: Wave One attend?} \\
    
    {\textbf{Response}: The singer of The Search for Everything: Wave One is John Mayer. John Mayer attended the Berklee College of Music in Boston, Massachusetts. So, the answer is \colorbox{green}{Berklee College of Music}.} \\

  \bottomrule
\end{tabular}
}
  \caption{Examples of responses generated using our approach for multihop bridge questions.}
  \label{tab:multihop_responses_examples}
\end{table*}

In this study, we show that our approach can be adapted to improve the performance on multi-hop questions.
Table \ref{tab:multihop_data_examples} shows examples of these questions. 
Recall that our approach works by mitigating hallucination / incorrectness in the sentences generated by the model. 
Thus, if we can enable the model to answer these multi-hop questions step by step, then our active detection and mitigation approach can be applied to these steps, leading to correct predictions.
To this end, we prompt the model and provide in-context examples demonstrating it to answer a given multi-hop question step by step.
Table \ref{tab:multihop_prompt} (in Appendix) shows the prompt with in-context examples used for this purpose.
Specifically, for a new question, the model generates the answer in multiple steps (one step at a time) and for each step, we apply our technique in which we first identify the low probability concepts from the sentence, validate their correctness using web search results, mitigate the hallucination (if detected), and then proceed to generate the next step.
In our case study, we sample $50$ multi-hop bridge questions from the validation set of HotpotQA \cite{yang-etal-2018-hotpotqa} and evaluate the performance.

\begin{figure}
    \centering
    \includegraphics[width=6.5cm]{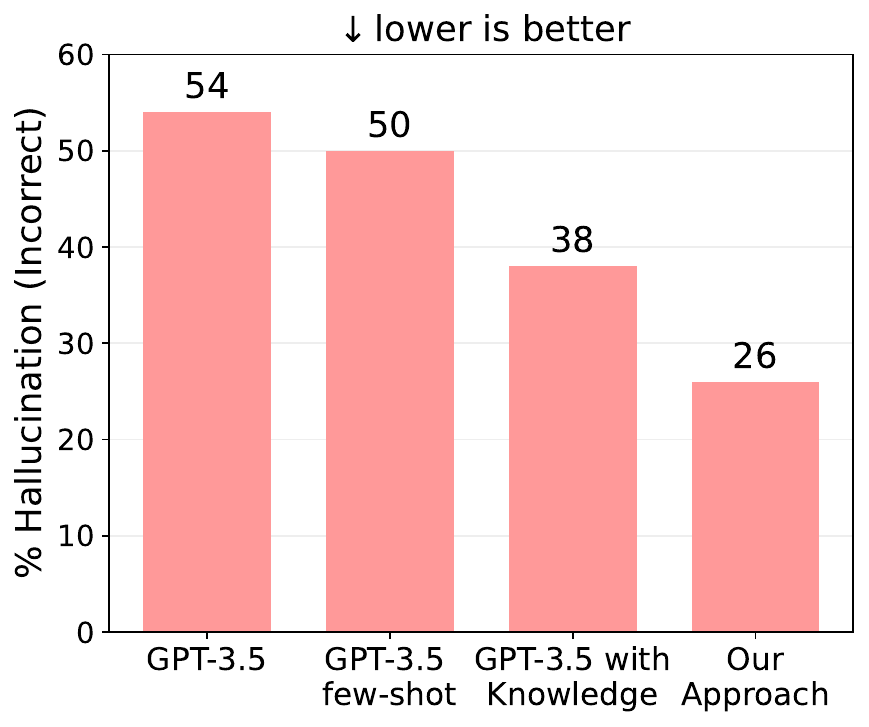}
    \caption{Comparing \% of hallucination on the Multi-hop Questions for GPT-3.5 model, GPT-3.5 with in-context examples, GPT-3.5 leveraging knowledge (retrieved via web search) and our approach. 
    }
    \label{fig:multihop_result}
\end{figure}

Table \ref{tab:multihop_responses_examples} shows examples of responses generated using our approach.
Figure \ref{fig:multihop_result} shows the performance achieved by different methods on this task.
First, it shows the performance of the GPT-3.5 model; the model answers $54\%$ of the questions incorrectly.
Then, it shows the performance of the GPT-3.5 model with in-context examples; it results in a slight improvement over the zero-shot performance.
Then, it shows the performance of the model on leveraging the knowledge retrieved from the web (using the question as the search query) as context to answer the question. 
As expected, the model's performance improves, i.e., it results in fewer incorrect predictions. 
Finally, we show the performance of our active detection and mitigation approach which results in considerably fewer hallucinations (26\%), i.e., higher percentage of correct answers.
This demonstrates the effectiveness of our approach in improving the performance on multi-hop questions.

\subsection{False Premise Questions}
\label{case_study_false_premise_questions}

\paragraph{Motivation and Experimental Setup:}
We perform this experiment because LLMs have already been shown to perform remarkably well on a wide range of `correct' questions, i.e., questions that are factually correct and make the right assumptions \cite{khashabi-etal-2020-unifiedqa,brown2020language,zhang2022opt,lourie2021unicorn,chowdhery2022palm,rae2021scaling}.
However, users in real-world applications often ask questions that are based on false premises / pre-suppositions such as ``Why energy is absorbed in exothermic reactions?'' and ``Why do floppy disks have higher storage capacity than USB drives?''.

We observe that state-of-the-art models often struggle to appropriately respond to such questions; thus, such questions serve as another challenging evaluation setting. 
To this end, we conduct a case study and compile a set of $50$ such adversarial questions, i.e., we compile questions on which GPT-3.5 model generates an incorrect response. 
This is done to create a challenging experimental setup for evaluation as the model generates incorrect output for such questions.
Furthermore, we also create a true premise question corresponding to each false premise question.
Table \ref{tab:adversarial_data_examples} (Appendix) shows examples of these question pairs.
For this task, we evaluate correctness at the complete answer level as this is a question answering task and the entire answer needs to be correct for the answer to be considered as correct.

\begin{figure}
    \centering
    \includegraphics[width=5.2cm]{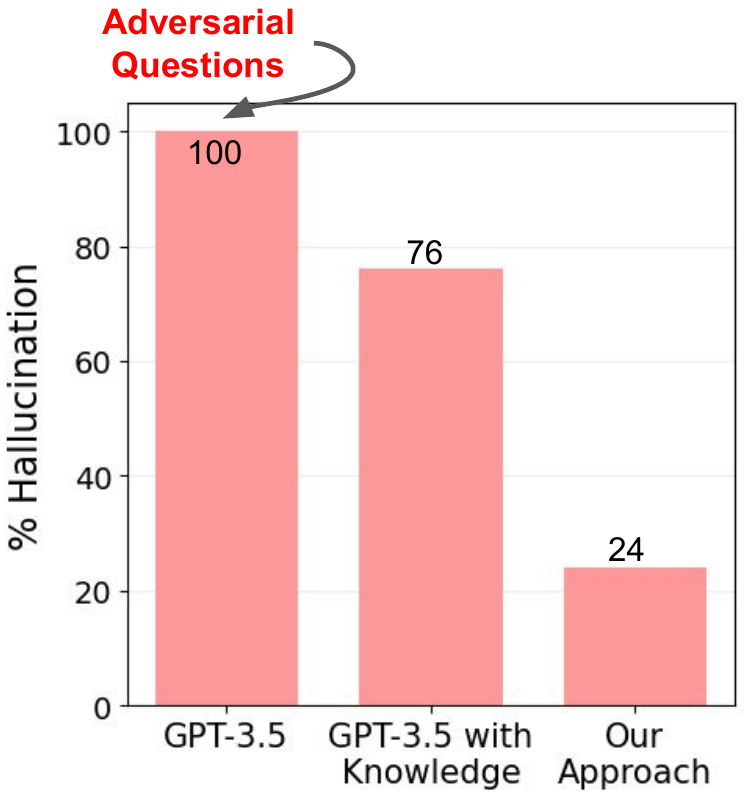}
    \caption{Comparing \% of hallucination on the `False Premise Questions' for GPT-3.5 model, GPT-3.5 leveraging knowledge (retrieved via web search) and our approach with question rectification (lower is better). 
    }
    \label{fig:adversarial_data_result}
\end{figure}

\paragraph{Approach:}
We note that an ideal response from a system for such questions depends on the application.
For instance, some applications may require identifying such questions and then abstaining on them like the selective prediction systems \cite{varshney-baral-2023-post, kamath-etal-2020-selective, xin-etal-2021-art, varshney-etal-2022-investigating}.
Some applications may additionally require suggesting a `rectified' question and providing response to that rectified question such as the search engines.
Our approach supports these requirements by using the validation and mitigation step on the given input question.

\begin{table*}[!htbp]
\centering
\small
{
\begin{tabular}
{@{}p{0.42\linewidth}p{0.56\linewidth}}
\toprule

  \textbf{Original Question} & \textbf{After Modification} \\ 
  \midrule

    \multicolumn{1}{r}{\textbf{\underline{False Premise Questions}}} \\ \\

    Why does Mars have three moons? & Why does Mars have two moons? (\cmark)\\

    Why are golf balls bigger than basketballs?	& Why are golf balls smaller than basketballs? (\cmark)\\

    What are some papers on the relationship between homeschooling and neuroplasticity?	& What are some papers on the relationship between homeschooling and learning outcomes? (\cmark)\\

    Why USA has the lowest happiness index? & What factors have contributed to the decline in happiness among Americans? (\cmark)\\

    How many metres does a typical apple weigh? & How many grams or ounces does a typical apple weigh? (\cmark)\\

    Why do gases have a particular shape? & Why do gases not have a definite volume or shape? (\cmark)\\

    Why do migrant workers never leave their home?	& Why do migrant workers leave their home? (\cmark)\\

    When a diver swims deeper, why does the water pressure declines? &	When a diver swims deeper, why does the water pressure increase? (\cmark)\\

    Why does Mars have higher gravity than Earth?	& Why does Mars have weaker gravity than Earth? (\cmark)\\


    Why do all rabbits have red eyes? & Why do some rabbits have red eyes? (\cmark)\\

    Why does Helium have atomic number of 1? & Why does Helium have atomic number of 2? (\cmark)\\ 

    Why does Bangladesh have the highest population in the world?  & Why does Bangladesh have the highest population growth rate in the world? (\xmark) \\

    Why are tigers' eggs bigger than chicken's eggs? & Why do some breeds of chickens lay larger eggs than others? (\xmark) \\

    \midrule

    \multicolumn{1}{r}{\textbf{\underline{True Premise Questions}}} \\  \\

    Why gases are shapeless? & 
    Why are gases shapeless? (\cmark)\\
    
    How did USA become a developed country? & 
    How did the United States become a developed country (\cmark)\\
    
    Why Afghanistan has a low happiness index? & 
    What factors contribute to Afghanistan's low happiness index? (\cmark)\\
    
    Why are golf balls smaller than basketballs? & 
    Why are golf balls typically smaller than basketballs? (\cmark)\\

    
    How were the 2020 USA presidential election? & 
    What were the results of the 2020 USA presidential election? (\cmark)\\
    
  \bottomrule
\end{tabular}
}
  \caption{Examples of original questions and the questions resulting from the rectification step. Our approach successfully modifies 76\% false premise questions and also modifies (without changing the semantics) $6$ instances of true premise questions.  \xmark and \cmark indicate that the modified question is incorrect and correct, respectively.}
  \label{tab:adversarial_data_rectification_examples}
\end{table*}

Specifically, we first retrieve the relevant knowledge (via Bing Search using the question as query). 
Then, conditioned on the retrieved knowledge, we prompt the model to respond `Yes' if the question makes factually correct assumptions, otherwise respond `No'.
If the response to this prompt is No, then we proceed to modify the question using the mitigation step. 
Table \ref{tab:false_premise_prompts} shows both the instructional prompts used for identifying and rectifying a potentially false premise question.
This step enables identifying false premise questions and also rectify them to facilitate the system in providing an appropriate response.
Importantly, we also show that our approach does not incorrectly modify a true premise question. This is crucial because if the user's question is correct then the system's response must be pertinent to that and not to its modified variant.

\paragraph{Performance Analysis:}

Recall that the false premise questions in our evaluation set are adversarially collected, i.e., GPT-3.5 gives an incorrect response to all of these questions.
First, we evaluate the performance of GPT-3.5 model when relevant knowledge (retrieved via bing search using the question as the search query) is given as context to answer the question.
We find that even with the retrieved knowledge, GPT-3.5 manages to answer only $24\%$ false premise questions correctly, i.e., hallucinates on the remaining $76\%$ questions.
In contrast, \textbf{our approach answers $\bm{76\%}$ questions correctly and hallucinates only on $\bm{24\%}$.}
Figure \ref{fig:adversarial_data_result} shows this comparison. 
Furthermore, we note that even in some of these $24\%$ hallucinated responses, some of the individual sentences in the responses are correct. 
However, since we focus on complete answer correctness, we mark them as incorrect.
Table \ref{tab:adversarial_response_examples} shows responses on a few false premise questions generated by the GPT-3.5 model, GPT-3.5 model leveraging the retrieved knowledge as context, and our approach.

\paragraph{Efficacy of Question Rectification:}
We analyze the performance of our approach in rectifying the questions; \textbf{it successfully repairs $\bm{76\%}$ false premise questions while not incorrectly modifying any true premise question.}
Though this step makes modifications in a small number of true premise questions ($6$ instances), it does not change the semantics of those questions as shown in the Table \ref{tab:adversarial_data_rectification_examples}.
We note that not incorrectly modifying a true premise question is an important characteristic of our approach.


\subsection{Other Applications}
Our approach has utility in a variety of other applications also such as {Abstractive Summarization} and {Claim Verification}. 
In abstractive summarization where the generated summary has been shown to be often hallucinated \cite{cao-etal-2022-hallucinated,zhao-etal-2020-reducing,chen-etal-2021-improving} can be improved using our approach. Note that in the validation procedure of our approach, the relevant knowledge for this task will be retrieved from the original document instead of the web. However, for open-summarization, knowledge can be additionally retrieved from the web also.
Our approach can be adapted for the claim verification task also as we can first identify the key sub-claims and then verify each sub-claim using the validation procedure. Here, the mitigation step will be useful for providing explanation behind the model's decision.
We leave exploring these other usecases of our approach for future work.

\section{Related Work}
Advancements in the field of natural language processing led to the development of models that possess an impressive ability to generate fluent and coherent text. However, these models are vulnerable to a phenomenon called text hallucination.
Prior work \cite{maynez-etal-2020-faithfulness,huang2021factual, ji2023survey} has categorized text hallucinations into two classes: Intrinsic (when the generated output contradicts the source content) and Extrinsic (when the generated output cannot be verified from the source content, i.e., it that can neither be supported nor contradicted by the source). 

One thread of research pertaining to hallucinations has focused on studying different causes of this phenomenon such as training data quality \cite{wang-2019-revisiting,lee-etal-2022-deduplicating}, source-target divergence \cite{dhingra-etal-2019-handling}, ill-suited modeling \cite{aralikatte-etal-2021-focus,feng2020modeling,li-etal-2018-ensure}, and randomness during inference \cite{dziri-etal-2021-neural, tian2019sticking, lee2022factuality}.

The other thread focuses on addressing the hallucination problem \cite{manakul2023selfcheckgpt, azaria2023internal, lee2022factuality, du2023improving, zhang2023interpretable}.
\citet{manakul2023selfcheckgpt} propose a sampling-based hallucination detection approach in which they first sample multiple responses from the model and then measure the information consistency between the different responses. They posit that when a language model knows a given concept well, the sampled responses are likely to be similar and contain consistent facts; on the other hand, for hallucinated facts, stochastically sampled responses are likely to diverge and may completely contradict one another.  

Another recent work \citet{azaria2023internal} leverages LLM's internal state to identify the truthfulness of a statement. Using an annotated dataset, they train a separate classifier that takes the LLM's activation values as input and predicts its truthfulness.
\citet{lee2022factuality} hypothesize that the randomness of sampling is more harmful to factuality when it is used to generate the latter part of a sentence than the beginning of a sentence and propose a new sampling algorithm named factual-nucleus sampling that dynamically adapts the `nucleus' p along the generation of each sentence.
\citet{du2023improving} propose an approach motivated by \textit{The Society of Mind} and \textit{multi-agent settings} in which multiple models individually propose and jointly debate their responses and reasoning processes to arrive at a common answer.
In our approach, we leverage the logit output values, web search, and actively detect and mitigate hallucinations.
We demonstrate the effectiveness of our approach on a variety of tasks, including article generation, multi-hop question answering, and false premise question answering.

\section{Conclusion}
In this work, we proposed an approach that actively `detects' and `mitigates' hallucinations of the large language models. 
Through systematic and extensive experiments with the article generation task, we showed that our approach successfully reduces the hallucinations of the GPT-3.5 (text-davinci-003) from $47.5\%$ to $14.5\%$ on average.
We also demonstrated the individual efficacy of our detection and mitigation techniques.
Specifically, our detection technique achieves a high recall and the mitigation technique successfully mitigates a large fraction of the correctly detected hallucinations.
Notably, the mitigation technique does not introduce new hallucinations even in the case of incorrectly detected hallucinations, i.e., false positives. 
We further demonstrated the effectiveness and wide applicability of our approach and presented several interesting studies including evaluation with another LLM (Vicuna) and answering multi-hop and false premise questions.
Overall, our work addresses the LLMs' hallucination problem and thus contributes to improving their reliability and trustworthiness, a crucial step en route to enabling their widespread adoption in real-world applications.

\bibliography{anthology,custom}
\bibliographystyle{acl_natbib}

\newpage
\appendix
\section*{Appendix}

\section{Fruther Details of the Approach}

Table \ref{tab:prompts} shows the instructional prompts used for different steps of our approach.
We note that these techniques are the preferred techniques as they do not require calling an external task-specific tool to achieve the corresponding objectives.

\subsection{Identify Key Concepts}
\label{examples_keywords}
Table \ref{tab:keyword_extraction_examples} shows examples of concepts identified using the three methods, i.e., Entity Extraction, Keyword Extraction, and Instructing the Model.
It shows that the entity extraction model misses many important concepts while the keyword extraction model identifies a lot of insignificant concepts also.
In contract, instruction technique successfully identifies majority of the important concepts.

\subsection{Create Validation Question}
\label{examples_validation_question}
Table \ref{tab:validation_question_examples} shows examples of validation questions corresponding to each concept created via instructing the model technique.
It shows examples of both the question types, i.e., Yes/No and Wh questions.
We prefer Yes/No questions as it is relatively easier to check the answer for these questions.
We leave exploring Wh-questions for validation for future work.

\section{Evaluation Data}

Table \ref{tab:sentence_statistics} shows the statistics of the sentences generated by the GPT-3.5 (text-davinci-003 with temperature 0) model. 
A sentence has $\sim18$ words on average and each sentence has $\sim3.2$ key concepts that are identified by our instruction technique.
\begin{table}[h]
    \centering
    \small
    \begin{tabular}{@{}lc@{}}
        \toprule
        \textbf{Statistic} & \textbf{Mean $\pm$ Std} \\
        \midrule
        \# Words in a Sentence & $18.6 \pm 5.55$ \\
        \# Key Concepts in a Sentence & $3.27 \pm$ 1.63 \\
        \# Words in a Key Concept & $1.79 \pm 1.02$ \\
    \bottomrule
    \end{tabular}
    \caption{Statistics of generated sentences.}
    \label{tab:sentence_statistics}
\end{table}

Table \ref{tab:sentence_annotation_example} shows examples of sentence-level and concept-level hallucination annotations.
\begin{table*}[]
\centering
\small
{
\begin{tabular}
{@{}cp{0.69\linewidth}}
\toprule

  \textbf{Step} & \textbf{Prompt} \\ 
  \midrule

    Input Prompt & \texttt{Write an article about \{topic\}} \\ \\

   {Identify Important Concepts} &	\texttt{Identify all the important keyphrases from the above sentence and return a comma separated list}. \\ \\
  
    {Create Validation Question}	& \texttt{For the above sentence about \{topic\}, generate a yes/no question that tests the correctness of \{concept\}.}  \\ \\
    
    {Answer Validation Question} &	\texttt{\{search results\} Answer the below question about {topic} in Yes or No based on the above context. \{validation question\}}. \\ \\
    
    {Repair Hallucinated Sentence} &	\texttt{The above sentence has information that can not be verified from the provided evidence, repair that incorrect information and create a new sentence based on the provided evidence.}\\ \\




  \bottomrule
\end{tabular}
}
  \caption{Instructional Prompts corresponding to different steps of our approach.}
  \label{tab:prompts}
\end{table*}

\begin{table*}[t]
\small
    \centering
    \resizebox{\linewidth}{!}{
    \begin{tabular}
    {@{}p{0.4\linewidth}|p{0.15\linewidth}p{0.18\linewidth}p{0.18\linewidth}}
    \toprule
        \textbf{Text} &
        \textbf{Entity Extraction} &
        \textbf{Keyword Extraction} & 
        \textbf{Instructing Model} \\
    \midrule
     
        John Russell Reynolds was an English physician and neurologist who made significant contributions to the field of neurology.
        & 
        
        John Russell Reynolds, English 
        &

         John Russell Reynolds,  English,  physician,  neurologist,  significant contributions,  field,  neurology

        & 
        John Russell Reynolds,  English,  physician,  neurologist,  neurology \\
        
        He was born in London in 1820 and studied medicine at the University of London. 
        &
        London, 1820, the University of London
        & 
        born,  London,  1820,  studied medicine,  University,  London
        & 
        London,  1820,  medicine,  University of London \\ \\

        After college, he worked as a lawyer for the PGA Tour, eventually becoming the Tour's Deputy Commissioner in 1989. 
        & 
        the PGA Tour, Tour, 1989
        &
        college,  worked,  lawyer,  PGA,  Tour,  eventually,  Tour,  Deputy Commissioner
        & 
        college,  lawyer,  PGA Tour,  Deputy Commissioner,  1989 \\ \\

        He was born in Sydney in 1971 and grew up in the city's western suburbs. 
        & 
        Sydney, 1971 
        & 
        born,  Sydney,  1971,  grew,  city,  suburbs
        & 
        Sydney,  1971,  western suburbs \\
       
    \bottomrule

    \end{tabular}
    }
    \caption{
    Examples of concepts identified by different techniques.
    }
    \label{tab:keyword_extraction_examples}
\end{table*}

\begin{table*}[t]
\small
    \centering
    \resizebox{\linewidth}{!}{
    \begin{tabular}
    {@{}p{0.1\linewidth}p{0.2\linewidth}p{0.1\linewidth}p{0.45\linewidth}}
    \toprule
        \textbf{Input} &
        \textbf{Generated Sentence} &
        \textbf{Concept} &
        \textbf{Validation Question} \\
    \midrule
        
        \multirow{15}{45pt}{Write an article about John Russell Reynolds}
        
         &
        
        \multirow{15}{85pt}{Reynolds was born in \textbf{London} in \textbf{1820} and studied \textbf{medicine} at the \textbf{University of London}.} 
        
        &
        \multirow{2}{17pt}{\textbf{London}}
        & 
         [Y/N] Was John Russell Reynolds born in London? 
         
         [Wh] Where was John Russell Reynolds born? \\ \\

        & 
        & 
        
        \multirow{2}{17pt}{\textbf{1820}} & 
        [Y/N] Was John Russell Reynolds born in 1820? 
        
        [Wh] What year was John Russell Reynolds born? \\ \\

        & 
        & 
        
        \multirow{3}{17pt}{\textbf{medicine}} & 
        [Y/N] Did John Russell Reynolds study medicine? 

        [Wh] What did John Russell Reynolds study at the University of London? \\ \\

        & 
        & 
        
        \multirow{4}{40pt}{\textbf{University of London}} & 
        [Y/N] Did Reynolds study medicine at the University of London? 

        [Wh] What university did John Russell Reynolds study medicine at? \\

    \bottomrule

    \end{tabular}
    }
    \caption{
    Examples of validation questions corresponding to the identified keyphrases generated by Instructing the Model technique.
    }
    \label{tab:validation_question_examples}
\end{table*}

\begin{table*}[]
\centering
\small
\resizebox{0.98\linewidth}{!}
{
\begin{tabular}
{@{}p{0.12\linewidth}p{0.58\linewidth}p{0.25\linewidth}}
\toprule

  \textbf{Sentence \#} & \textbf{Sentence} & \textbf{Sentence-level Correctness} \\ 
  \midrule

    \textbf{Sentence 1} & Eleanor Arnason is an \colorbox{green}{award-winning} \colorbox{green}{science fiction} and \colorbox{green}{fantasy} author who has been writing since the \colorbox{green}{1970s}. & Correct \\
    
    \textbf{Sentence 2} & She is best known for her novel \colorbox{green}{A Woman of the Iron People}, which won the \colorbox{green}{James Tiptree Jr. Award} in \colorbox{green}{1991}. & Correct \\

    \textbf{Sentence 3} & Her work has been praised for its exploration of \colorbox{green}{gender}, \colorbox{green}{race}, and \colorbox{green}{identity}, as well as its \colorbox{green}{imaginative world-building}. & Correct \\ 

    \textbf{Sentence 4} & Arnason was born in \colorbox{red}{Minneapolis}, \colorbox{red}{Minnesota} in \colorbox{green}{1942}. & Hallucination \\

    \textbf{Sentence 5} & She attended the \colorbox{green}{University of Minnesota}, where she earned a \colorbox{red}{degree} in \colorbox{red}{English literature}. & Hallucination \\

 \bottomrule
\end{tabular}
}
  \caption{Examples of both sentence and concept-level annotations for the input: ``Write an article about Eleanor Arnason''. 
  Annotation for correct concepts is represented in \colorbox{green}{green} while annotation for hallucinated concept is represented in \colorbox{red}{red}.}
  \label{tab:sentence_annotation_example}
\end{table*}

\section{Recall of Hallucination Detection vs Probability Threshold}

\begin{figure*}[t]
\centering
    \begin{subfigure}{.45\textwidth}
    \caption{Sentence-level}
        \includegraphics[width=\linewidth]{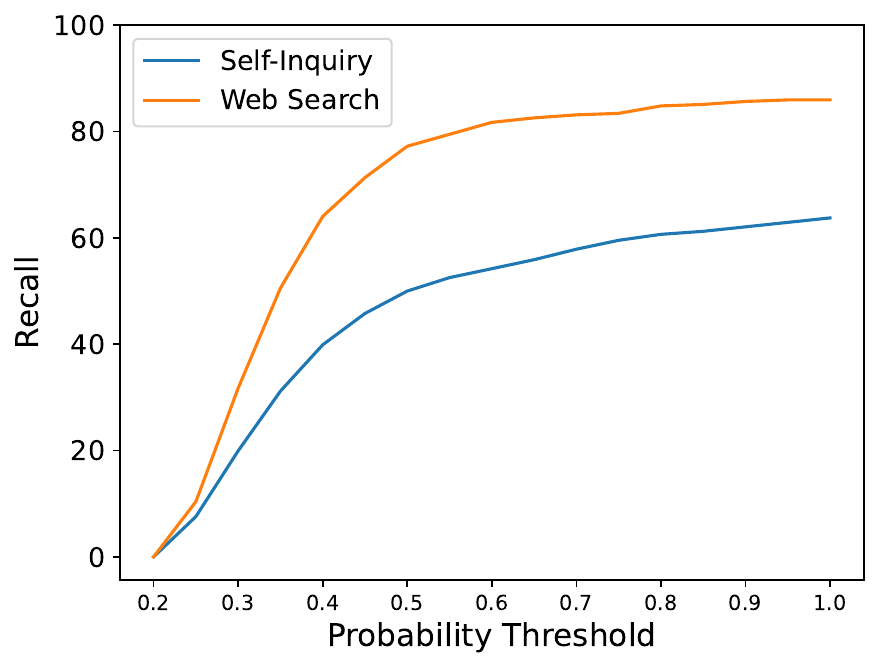}
        
    \end{subfigure} 
    \begin{subfigure}{.45\textwidth}
    \caption{Concept-level}
        \includegraphics[width=\linewidth]{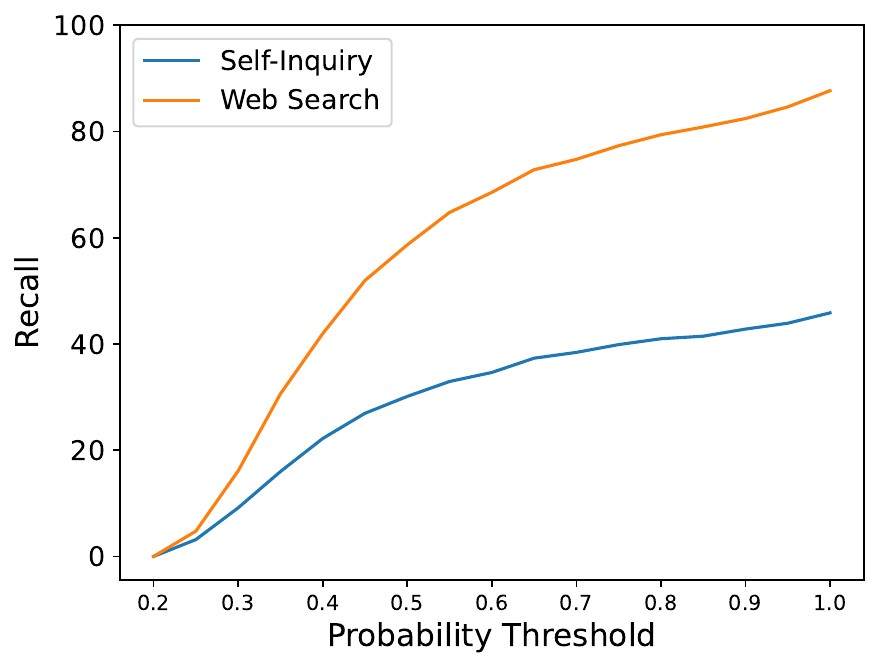}
        
    \end{subfigure}
    
    \caption{Recall of hallucination detection vs Probability threshold plot for Self Inquiry and web search techniques at both sentence-level and concept-level.
    }
    \label{fig:recall_vs_prob_self_inquiry_and_web_search_comparison}    
\end{figure*}

Figure \ref{fig:recall_vs_prob_self_inquiry_and_web_search_comparison} compares recall of hallucination detection for self-inquiry and web search techniques at different probability thresholds.
Web search considerably outperforms self-inquiry at all thresholds.

\section{Hallucination Mitigation Examples}

\begin{table*}[]
\centering
\small
{
\begin{tabular}
{@{}p{0.18\linewidth}p{0.8\linewidth}}
\toprule

     \textbf{Topic} & Rick Mahler \\
     
     \textbf{Original Sentence} & Mahler was born in \colorbox{red}{San Diego, California} in \colorbox{red}{1956}. (\xmark)	\\

     \textbf{Hallucination Reason} & He was born in Austin in 1953. \\
     
     \textbf{Modified Sentence} & Rick Mahler was born in Austin, Texas on August 5, 1953. (\cmark) \\ 
     \midrule

     \textbf{Topic} & Matthew Aylmer, 1st Baron Aylmer \\
     
     \textbf{Original Sentence} & Baron Aylmer, was an English naval officer and politician who served as a Member of Parliament and later as a peer in the \colorbox{red}{House of Lords}. (\xmark)	\\

     \textbf{Hallucination Reason} & He was in House of Commons not House of Lords. \\
     
     \textbf{Modified Sentence} & Baron Aylmer was an Anglo-Irish Royal Navy officer and Whig politician who sat in the English and British House of Commons between 1695 and 1720. (\cmark)\\ 
     \midrule

     \textbf{Topic} & Derek King\\
     
     \textbf{Original Sentence} & In 2014, Derek King released his debut single \colorbox{red}{"Let's Go,"} which quickly became a hit in Canada. (\xmark)	\\

     \textbf{Hallucination Reason} & "Let's Go" was not his song. \\
     
     \textbf{Modified Sentence} & In 2014, Derek King released his debut mixtape Fake I.D., which featured fellow Bay Area artists Rayven Justice and IAMSU!, among others. (\cmark)\\ 
     \midrule

     \textbf{Topic} & Marshall Manesh\\
     
     \textbf{Original Sentence} & Marshall Manesh is an Iranian-American actor best known for his roles on the television shows How I Met Your Mother and \colorbox{red}{The Middle}. (\xmark)	\\

     \textbf{Hallucination Reason} & He was not in The Middle. \\
     
     \textbf{Modified Sentence} & Marshall Manesh is an Iranian-American actor best known for his recurring roles on the television shows Will \& Grace, Scrubs, Andy Barker, P.I., Hot in Cleveland, Boston Legal, and How I Met Your Mother, where he played taxi driver Ranjit. (\cmark)\\ 
     \midrule

     

     

     \textbf{Topic} & William J. Flanagan, Jr. \\
     
     \textbf{Original Sentence} &   He is the \colorbox{red}{founder and CEO of Flanagan Financial Group, a financial services firm based in} \colorbox{red}{New York City}. (\xmark)	\\

     \textbf{Hallucination Reason} & The complete sentence is hallucinated as he is a retired US Navy admiral.\\
     
     \textbf{Modified Sentence} & He is a retired four-star admiral who served as Commander in Chief, United States Atlantic Fleet from 1994 to 1996 and is the recipient of numerous military awards. (\cmark)\\






  \bottomrule
\end{tabular}
}
  \caption{Examples of \textbf{successful} mitigation of hallucinations by our mitigation technique. 
  Original Sentence corresponds to the sentence generated by the model and Modified Sentence corresponds to the sentence obtained on applying our technique.}
  \label{tab:mitigation_examples}
\end{table*}

\begin{table*}[]
\centering
\small
{
\begin{tabular}
{@{}p{0.18\linewidth}p{0.8\linewidth}}
\toprule

     \textbf{Topic} & Matthew Aylmer, 1st Baron Aylmer \\
     
     \textbf{Original Sentence} & Aylmer joined the Royal Navy in \colorbox{red}{1790} and served in the \colorbox{red}{French Revolutionary Wars} and \colorbox{red}{the Napoleonic Wars}. (\xmark)	\\

     \textbf{Hallucination Reason} & He did not serve in these wars. \\
     
     \textbf{Modified Sentence} & Aylmer entered the Royal Navy under the protection of the Duke of Buckingham as a Lieutenant in 1678 and served in the \colorbox{red}{French Revolutionary Wars and the Napoleonic Wars}. (\xmark)\\ 

     \textbf{Hallucination Reason} & It rectified the date but failed to correct the hallucination about wars. \\     
     
     \midrule

     \textbf{Topic} & K. S. Manilal \\
     
     \textbf{Original Sentence} &   Manilal was a prolific writer and translator, having written more than \colorbox{red}{50 books} and translated over 100 works from \colorbox{red}{English, Sanskrit}, and other languages into Malayalam. (\xmark)	\\

     \textbf{Hallucination Reason} & He has not written 50 books and translated works to English and Malayalam.\\
     
     \textbf{Modified Sentence} & Manilal was a prolific researcher and translator, having translated Hendrik van Rheede's 17th century Latin botanical treatise, Hortus Malabaricus, into English, \colorbox{red}{Sanskrit}, and Malayalam. (\xmark)\\ 

     \textbf{Hallucination Reason} & The information about Hortus Malabaricus is correct but he translated it into English and Malayalam only and not Sanskrit. \\     
     
     \midrule

     \textbf{Topic} & Freddie Frith \\
     
     \textbf{Original Sentence} &  He was born in \colorbox{red}{London} in \colorbox{red}{1929} and began his racing career in \colorbox{red}{1951}. (\xmark)	\\

     \textbf{Hallucination Reason} & He was born in Grimsby in 1909 and began his career in 1930. \\
     
     \textbf{Modified Sentence} & Freddie Frith was born in Grimsby, England in 1909 and began his racing career in \colorbox{red}{1935} when he won the Junior Manx Grand Prix. (\xmark)\\ 

     \textbf{Hallucination Reason} & He did win the Junior Grand Prix in 1935 but he began hi career in 1930. \\     
     
     \midrule

     \textbf{Topic} & Harry S. Kennedy \\
     
     \textbf{Original Sentence} &   Harry S. Kennedy was an \colorbox{red}{American politician who served as the 35th President of the United States} \colorbox{red}{from 1961 to 1963}. (\xmark)	\\

     \textbf{Hallucination Reason} & This sentence is true for John F. Kennedy not Harry S. Kennedy.\\
     
     \textbf{Modified Sentence} & John F. Kennedy was an American politician who served as the 35th President of the United States from 1961 to 1963. (\xmark)\\ 

     \textbf{Hallucination Reason} & This sentence is not hallucinated but it is not related to the topic. \\     
     

  \bottomrule
\end{tabular}
}
  \caption{Examples where our mitigation technique fails to mitigate complete hallucination in the generated sentence. 
  Original Sentence corresponds to the sentence generated by the model and Modified Sentence corresponds to the sentence obtained on applying our technique.}
  \label{tab:incorrect_mitigation_examples}
\end{table*}

Tables \ref{tab:mitigation_examples} and \ref{tab:incorrect_mitigation_examples} show examples where our mitigation technique successfully mitigates the hallucinations and where it fails, respectively.
We observe that in many of the failure cases, our technique fixes some hallucinated content of the sentences but fails to fix ALL the hallucinated content from them.
Furthermore, in some of the failure cases, our technique results in a sentence which is no longer hallucinated but it not completely related to the topic.

\section{Multi-hop QA Experiment}
\begin{table*}[]
\centering
\small
{
\begin{tabular}
{@{}p{0.74\linewidth}p{0.24\linewidth}}
\toprule

  \textbf{Question} & \textbf{Answer} \\ 
  \midrule

    The football manager who recruited David Beckham managed Manchester United during what timeframe? & from 1986 to 2013 \\

    The Vermont Catamounts men's soccer team currently competes in a conference that was formerly known as what from 1988 to 1996? & the North Atlantic Conference \\
    
    Ralph Hefferline was a psychology professor at a university that is located in what city? & New York City \\

    What is the county seat of the county where East Lempster, New Hampshire is located? & Newport \\

    Blackfin is a family of processors developed by the company that is headquartered in what city? & Norwood, Massachusetts \\

  \bottomrule
\end{tabular}
}
  \caption{Examples of  multihop questions from HotpotQA.}
  \label{tab:multihop_data_examples}
\end{table*}
\begin{table*}[]
\centering
\small
{
\begin{tabular}
{@{}p{0.99\linewidth}}
\toprule


    \texttt{\textbf{Question}: Which team does the player named 2015 Diamond Head Classic’s MVP play for?} \\
    
    \texttt{\textbf{Answer}: The 2015 Diamond Head Classic’s MVP was Buddy Hield. Buddy Hield plays for the Sacramento Kings of the NBA. So, the answer is Sacramento Kings.} \\ \\
    
    \texttt{\textbf{Question}: The Dutch-Belgian television series that "House of Anubis" was based on first aired in what year?}  \\
    
    \texttt{\textbf{Answer}: "House of Anubis" was based on is Het Huis Anubis. Het Huis Anubis was first aired in 2006. So, the answer is 2006.} \\ \\ 
    
    \texttt{\textbf{Question}: What is the birth date of the coach of the 2014-15 Michigan Wolverines men's basketball team?} \\
    
    \texttt{\textbf{Answer}: The coach of the 2014-15 Michigan Wolverines men's basketball team was John Beilein. John Beilein's birth date is February 5, 1953.} \\ \\
    
    \texttt{\textbf{Question}: \{question\} } \\
    \textbf{Answer}: \\

  \bottomrule
\end{tabular}
}
  \caption{Prompt with in-context examples used for prompting the model to answer `bridge' multihop questions step by step.}
  \label{tab:multihop_prompt}
\end{table*}

Table \ref{tab:multihop_data_examples} shows examples of multi-hop bridge questions from HotpotQA dataset.
Table \ref{tab:multihop_prompt} shows the prompt with in-context examples used for prompting the model to answer multi-hop questions step by step.
Table \ref{tab:multihop_responses_examples} shows examples of responses generated using our approach for multi-hop bridge questions.

\section{False Premise QA Experiment}
\begin{table*}[]
\centering
\small
{
\begin{tabular}
{@{}p{0.48\linewidth}p{0.48\linewidth}}
\toprule

  \textbf{False Premise Question} & \textbf{True Premise Question} \\ 
  \midrule

    Why does Mars have \textbf{three} moons?	& Why does Mars have two moons?\\

    Why are golf balls \textbf{bigger} than basketballs?	& Why are golf balls smaller than basketballs? \\

    What are some papers on the relationship between homeschooling and \textbf{neuroplasticity}?	& What are some papers on the relationship between homeschooling and college preparedness? \\

    Why do migrant workers \textbf{never leave} their home?	& Why do migrant workers leave their home? \\

    When a diver swims deeper, why does the water \textbf{pressure declines}? &	When a diver swims deeper, why does the water pressure increases? \\

    Why does Mars have \textbf{higher gravity} than Earth?	& Why does Mars have lower gravity than Earth? \\

    Why is \textbf{baseball the most popular sport} in the world? & Why is soccer the most popular sport in the world? \\

    How does \textbf{Earth make up over 99\% of the solar system's mass}? & How does Sun make up over 99\% of the solar system's mass? \\

    Why does \textbf{Helium have atomic number of 1}? & Why does Hydrogen have atomic number of 1? \\

  \bottomrule
\end{tabular}
}
  \caption{Examples of `false premise' questions and their corresponding `true premise' counterparts.}
  \label{tab:adversarial_data_examples}
\end{table*}
\begin{table*}[]
\centering
\small
{
\begin{tabular}
{@{}cp{0.93\linewidth}}
\toprule

  \textbf{Step} & \textbf{Prompt} \\ 
  \midrule


  
    
    

    {I} &	\texttt{Context: \{search results\} Question: \{question\} Based on the above context, does the above question make factually correct assumptions?} \\  \\

    {II} &	\texttt{Context: \{response\_of\_Step\_I\} Question: \{question\} Rectify the incorrect information in the given question based on the context.} \\ 


  \bottomrule
\end{tabular}
}
  \caption{Instructional Prompts for rectifying the false premise questions.}
  \label{tab:false_premise_prompts}
\end{table*}

\begin{table*}[]
\centering
\small
{
\begin{tabular}
{@{}p{0.2\linewidth}p{0.78\linewidth}}
\toprule


  \textbf{False Premise Question} &  \textit{Why are golf balls \textbf{bigger} than basketballs?}	\\ \\

  \textbf{GPT-3.5} &

    \colorbox{red}{Golf balls are bigger than basketballs because} they need to be able to travel farther distances and have a larger surface area to create more spin and control. The larger size also helps to reduce air resistance, allowing the ball to travel farther.  (\xmark) \\

    \textbf{GPT-3.5 w/ Knowledge} & 

    \colorbox{red}{Golf balls are bigger than basketballs because} they need to be heavier and denser in order to travel farther when hit with the same amount of force. The larger size allows for more mass-density, which gives the ball more momentum and distance when hit. (\xmark)\\

    \textbf{Our Approach} & 

    \colorbox{green}{Golf balls are smaller than basketballs because} they are optimized for golfers to provide the most ball velocity and distance possible, along with a controllable spin rate. The diameter of a golf ball must reach 1.68 inches, or 42.67 mm, in diameter, and a weight of no more than 1.62 ounces, or 45. (\cmark)\\


    \midrule

  \textbf{False Premise Question} &  \textit{Why do deserts have \textbf{higher} soil moisture than grasslands?}	\\ \\

  \textbf{GPT-3.5} &

    \colorbox{red}{Deserts have higher soil moisture than grasslands because} they receive more rainfall. Deserts typically receive less rainfall than grasslands, but the rain that does fall is more concentrated and intense. This causes the soil to absorb more moisture than in grasslands, where the rainfall is more spread out and less intense. (\xmark) \\

    \textbf{GPT-3.5 w/ Knowledge} & 

    \colorbox{red}{Deserts have higher soil moisture than grasslands because} they receive more precipitation than grasslands. Deserts are located in areas where the demand for water by the atmosphere (evaporation) and plants (transpiration) is much greater than precipitation. This means that the soil in deserts is able to retain more moisture than the soil in grasslands. (\xmark) \\

    \textbf{Our Approach} & 

    \colorbox{green}{Deserts have lower soil moisture than grasslands because} they have less water vapor in the air, which makes it harder to trap heat or cold in a desert. Additionally, deserts typically have less precipitation than grasslands, which further contributes to lower soil moisture levels. (\cmark)\\

    \midrule

  \textbf{False Premise Question} &  \textit{Why is it more \textbf{difficult} to swim in seawater than fresh river water?}	\\ \\

  \textbf{GPT-3.5} &

    Seawater is more dense than fresh river water due to its higher salt content. This increased density makes it \colorbox{red}{more difficult to swim in because} it requires more effort to move through the water. Additionally, the salt content of seawater can cause skin irritation and dehydration, making it more uncomfortable to swim in. (\xmark) \\

    \textbf{GPT-3 w/ Knowledge} & 

    \colorbox{red}{It is more difficult to swim in seawater than fresh river water because} seawater is more dense than fresh river water. This means that the buoyancy of the swimmer is reduced in seawater, making it more difficult to stay afloat. Additionally, the salt content in seawater can cause skin irritation and dehydration, making it more difficult to swim. (\xmark) \\

    \textbf{Our Approach} & 

    \colorbox{green}{It is easier to swim in seawater than fresh river water because} seawater is denser than fresh water due to its higher mineral content, including sodium, chloride, sulphate, magnesium and calcium. This increased density provides more buoyancy, making it easier for swimmers to stay afloat. (\cmark)\\

  \bottomrule
\end{tabular}
}
  \caption{
  Comparing responses generated on a few false premise questions by the GPT-3.5 model, GPT-3.5 moel leveraging the retrieved knowledge as context, and our approach.
  }
  \label{tab:adversarial_response_examples}
\end{table*}

Table \ref{tab:adversarial_data_examples} shows examples of false premise and true premise question pairs.
Table \ref{tab:adversarial_response_examples} shows responses generated on a few false premise questions by the GPT-3.5 (text-davinci-003) model, GPT-3.5 (text-davinci-003) using the retrieved knowledge as context, and our approach.

\end{document}